\documentclass[10pt,twocolumn,letterpaper]{article}

\usepackage[pagenumbers]{cvpr}

\usepackage{times}
\usepackage{epsfig}
\usepackage{graphicx}
\usepackage{amsmath}
\usepackage{amssymb}

\usepackage{import}
\usepackage[dvipsnames]{xcolor}         %
\usepackage{textcomp}
\usepackage{booktabs}
\usepackage[super]{nth} %
\usepackage{xcolor,colortbl}
\usepackage{multirow}
\usepackage{rotating}
\usepackage{subcaption}
\usepackage{bm}
\usepackage{cases}

\definecolor{cvprblue}{rgb}{0.21,0.49,0.74}
\usepackage[pagebackref,breaklinks,colorlinks,citecolor=cvprblue]{hyperref}

\usepackage[capitalize]{cleveref}
\crefname{section}{Sec.}{Secs.}
\Crefname{section}{Section}{Sections}
\Crefname{table}{Table}{Tables}
\crefname{table}{Tab.}{Tabs.}

\newcommand{\SCR}{\textit{SCR}}
\newcommand{\SfM}{\textit{SfM}}
\newcommand{\PnP}{\textit{PnP}}
\newcommand{\ours}{SACReg\xspace}
\newcommand{\uline}[1]{\underline{#1}}

\newcommand{\I}{\mathcal{I}}
\newcommand{\R}{\mathcal{R}}
\newcommand{\C}{\mathcal{C}}
\newcommand{\Raug}{\mathcal{R}'}
\newcommand{\V}{\mathcal{V}}
\newcommand{\p}{\mathbf{p}}
\newcommand{\pv}{\mathbf{v}}
\newcommand{\x}{\mathbf{x}}
\newcommand{\y}{\hat{\mathbf{y}}}
\newcommand{\bd}[1]{\textbf{#1}}

\newcommand{\PAR}[1]{\noindent{\bf{#1}}}
\newcommand{\hl}[0]{\cellcolor{gray!22}} %

\newlength{\SCRfigwidth}
\newlength{\SCRfigheight}
\newcommand{\SCRraisedgraphics}[1]{\raisebox{-.5\height}[\dimexpr0.5\height+2pt]{\includegraphics[width=\SCRfigwidth, height=\SCRfigheight, keepaspectratio]{#1}}}
\newenvironment{SCRfigtabular}
{
    \setlength{\SCRfigwidth}{0.16\linewidth}
    \setlength{\SCRfigheight}{\SCRfigwidth}
    \setlength{\tabcolsep}{3pt}
    \bf{}
    \small{}
    \begin{tabular}{c@{ }c@{ }cp{0.0cm}cp{0.0cm}c@{ }c}
    \multicolumn{3}{c}{Reference images with sparse 2D/3D annotations} &
    & Query image &
    & Predicted coordinates
    & Confidence \\
}
{
    \end{tabular}
}
\newcommand{\SCRfigline}[1]
{
    \SCRraisedgraphics{#1/refpoints0.jpg} &
    \SCRraisedgraphics{#1/refpoints1.jpg} &
    \SCRraisedgraphics{#1/refpoints2.jpg} & &
    \SCRraisedgraphics{#1/query.jpg} & &
    \SCRraisedgraphics{#1/prediction.jpg} &
    \SCRraisedgraphics{#1/confidence.jpg}
}

\begin{document}

\title{\ours: Scene-Agnostic Coordinate Regression for Visual Localization}

\author{Jerome Revaud$^\dagger$
\qquad \qquad
Yohann Cabon$^\dagger$
\qquad \qquad
Romain Br\'egier$^\dagger$
\\
JongMin Lee$^*$
\qquad \qquad
Philippe Weinzaepfel$^\dagger$\\[0.2cm]
\quad\quad$^\dagger$Naver Labs Europe\quad\quad\quad\quad\quad\quad\quad\quad$^*$Seoul University\\[-0.1cm]
{\tt\small firstname.lastname@naverlabs.com}\quad\quad\quad\quad{\tt\small sdrjseka96@naver.com}
\vspace{-0.5cm}
}

\maketitle

\begin{abstract}
Scene coordinates regression (SCR), \ie, predicting 3D coordinates for every pixel of a given image, has recently shown promising potential.
However, existing methods remain %
limited to small scenes %
memorized
during training, and thus hardly scale to realistic datasets and scenarios.
In this paper, we propose a generalized SCR model trained once to be deployed in new test scenes, regardless of their scale, without any finetuning.
Instead of encoding the scene coordinates into the network weights, our model takes as input a database image with some sparse 2D pixel to 3D coordinate annotations, extracted from \eg off-the-shelf Structure-from-Motion or RGB-D data, and a query image for which are predicted a dense 3D coordinate map and its confidence, based on cross-attention.
At test time, we rely on existing off-the-shelf image retrieval systems and fuse the predictions from a shortlist of relevant database images \wrt the query. Afterwards camera pose is obtained using standard Perspective-n-Point (PnP).
Starting from self-supervised CroCo pretrained weights, we train our model on diverse datasets to ensure generalizabilty across various scenarios, and significantly outperform other scene regression approaches, including scene-specific models, on multiple visual localization benchmarks.
Finally, we show that the database representation of images and their 2D-3D annotations can be highly compressed with negligible loss of localization performance.
\end{abstract}

\begin{figure}
    \centering
    \includegraphics[trim=0 140 200 0,clip,width=\linewidth]{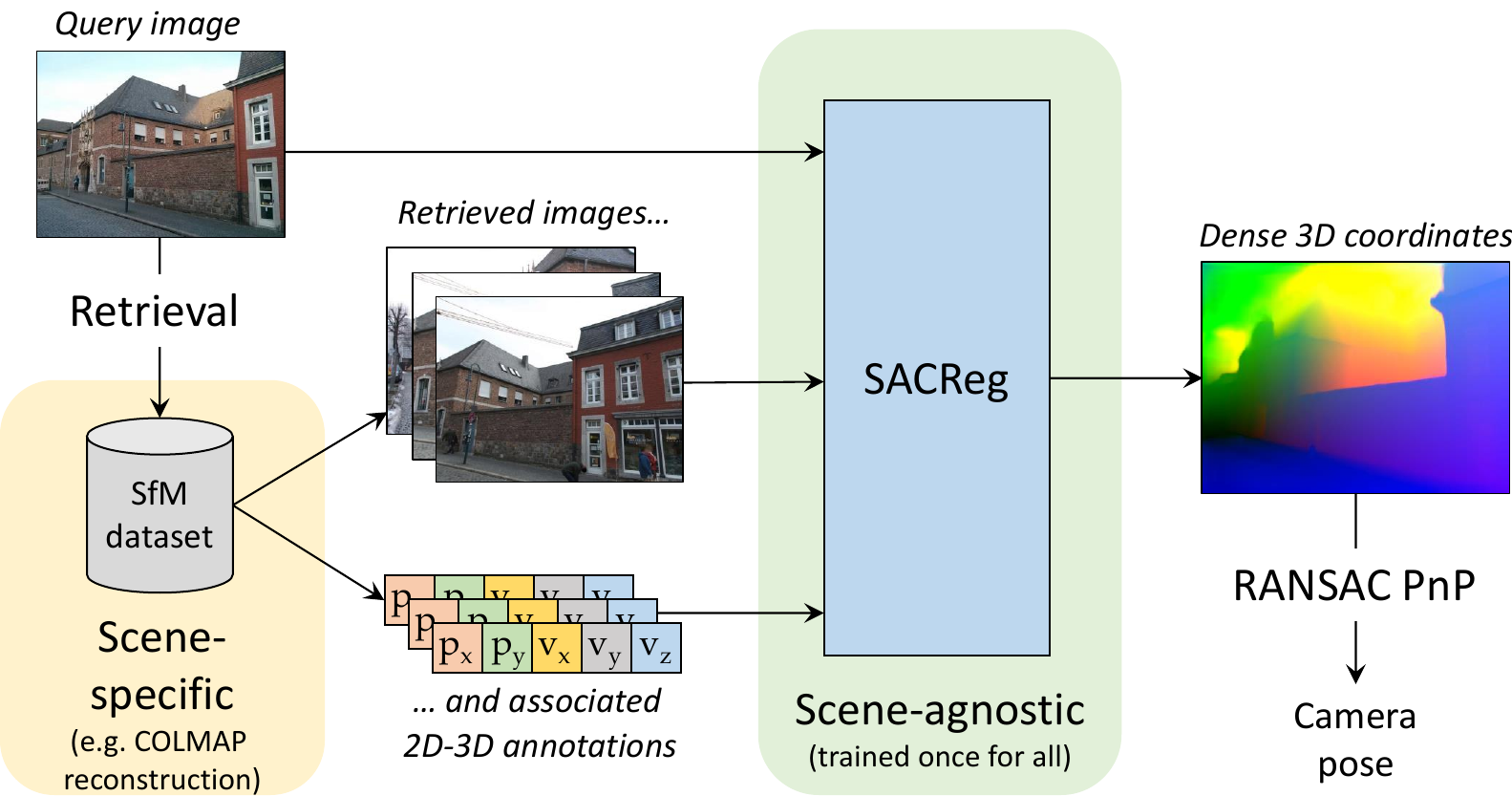} \\[-0.2cm]
    \caption{%
    \textbf{Scene-agnostic coordinate regression (\ours) for visual localization.}
    Given a query image and a set of related views with sparse 2D-3D annotations retrieved from a database (bottom left), \ours predicts absolute 3D coordinates for each pixel of the query image (right). These can be used for visual localization using a robust \PnP{} algorithm (bottom right). Importantly, \ours is scene-agnostic: %
    it can be used in any novel scene without re-training,
    only the images and 2D-3D annotations that serve as input are scene-specific.}
    \label{fig:overview}

    \vspace{-0.4cm}
    
\end{figure}

\section{Introduction}
\label{sec:intro}

Image-based scene coordinate regression (\SCR{}) consists in predicting the 3D coordinates of the point associated to each pixel of a given query image. %
\SCR{} methods have numerous applications in computer vision, and previous work has shown promising potential over the last few years.
Such methods have for instance been proposed for visual localization~\cite{shotton_scene_2013,yang_sanet_2019,tang_learning_2021,dsacstar} in combination with a Perspective-n-Point (\PnP{}) solver~\cite{pnp}. 
Other applications include object pose estimation~\cite{brachmann_learning_2014,zakharov_dpod_2019}, depth completion~\cite{penet, lidar_completion_uncertainty, cspnplus, nonlocal_spn,semattnet}, augmented reality or robotics~\cite{shotton_scene_2013,sc_wls_eccv2022}.

Unfortunately, existing \SCR{} approaches pose significant scalability issues and end up being rather impractical.
Most of the time, 3D scene coordinates are directly embedded into the parameters of the learned model, being it a random forest~\cite{shotton_scene_2013} or a neural network~\cite{dsac,dsacstar,esac,sc_wls_eccv2022}, hence \emph{de facto} limiting one model to a specific, generally small, scene for which it was trained on.
Some recent attempts to mitigate this issue, such as training different experts~\cite{esac},
sharing scene-agnostic knowledge between scenes~\cite{neumap}, or heavily relying on dense 3D reconstructions at test time~\cite{tang_learning_2021,yang_sanet_2019}, improve by some aspects but  still require scene-specific finetuning, %
can be limited to small scenes, and do not offer scene-agnostic solutions yet.
In essence, there is no \emph{universal} SCR model that can seamlessly function \emph{as-is} on any given test scene.

In this paper, we propose a new paradigm for scene coordinates regression that allows to train a generic model
once, and deploy it %
to novel scenes of arbitrary scale.
As illustrated in Figure~\ref{fig:overview}, %
our scene-agnostic coordinate regression (\ours) model takes as input a query image as well as a set of relevant database images for which 3D scene coordinates are available at sparse 2D locations.
\ours predicts dense 3D coordinates, for each pixel of the query image.
From this output, the camera pose can be obtained by solving a Perspective-n-Point (PnP) problem. 
Note that all inputs of \ours can be obtained via off-the-shelf methods: relevant database images can be obtained using image retrieval techniques~\cite{fire,kapture,apgem}, while the sparse 2D-3D correspondences are a by-product of map construction procedures, \ie, obtained using dedicated sensors %
or Structure-from-Motion (SfM) pipelines~\cite{SchonbergerCVPR16StructureFromMotionRevisited}. 

\begin{figure*}
    \centering
    \includegraphics[trim=0 280 0 0,clip,width=0.75\linewidth]{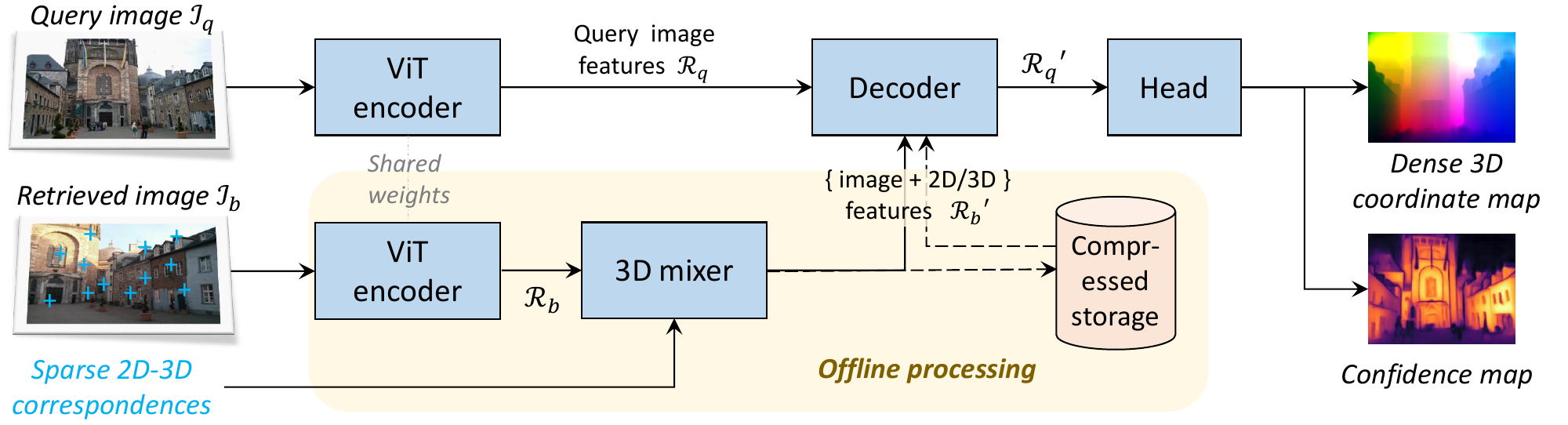} \\[-0.3cm]
    \caption{\textbf{Overview of the \ours architecture} for a given %
    pair of query and relevant database image.
    Both images are first encoded by a vision transformer, then sparse 2D-3D correspondences %
    are used to augment the encoded tokens of the database image with geo-spatial information. %
    A decoder jointly processes both sets of tokens and outputs a dense 3D coordinate map and an associated confidence map using a specific head. 
    Database images can be encoded offline with their 2D-3D annotations and compressed for better test-time efficiency. 
    }
    \label{fig:crocordinate}
    \vspace{-0.29cm}
\end{figure*}

In summary, our first contribution is to introduce a generic model for scene-agnostic coordinate regression. %
It uses a Vision Transformer (ViT)~\cite{vit} to encode query and a database image, as illustrated in Figure~\ref{fig:crocordinate}. 
Database image tokens are augmented with %
their provided sparse 2D-3D correspondences,
using a transformer decoder. %
Afterward, another transformer decoder combines these augmented tokens with those extracted from the query image, which are further processed by a convolutional head to regress dense 3D scene coordinates and an associated pixelwise confidence map.
Finally, predictions made separately for each database image are fused based on the confidence values.

As a second contribution, we propose to regress an encoding of the 3D coordinates rather than the raw 3D coordinates.
Doing so solves a major limitation of existing scene-agnostic approaches which assume small scenes with zero-centered coordinate systems and cannot generalize to unbounded scenes~\cite{tang_learning_2021,yang_sanet_2019}.
To that aim, we introduce an invertible and noise-resistant cosine-based encoding of 3D coordinates.
We show that it can generalize effortlessly to arbitrary coordinate ranges at test time. 

As a third contribution, we show that the augmented database tokens (combining image and associated 2D-3D correspondences) can be pre-computed and compactly stored for faster inference.
Specifically, using simple product quantization (PQ)~\cite{jegou_pq_2011}, %
we achieve compression rates over 30 for VGA images with no loss of performance, reducing the storage needs from 3.7MB to 115kB per image.
This simple scheme significantly outperforms recent compression approaches for visual localization and sets a new state of the art of database footprint.

Lastly,
we report on par or better performance than existing state-of-the-art SCR approaches on multiple benchmarks without any finetuning.
To ensure generalization, we initialize the network weights with cross-view completion pretraining (CroCo)~\cite{croco,crocostereo} and train on diverse sources: outdoor buildings with the MegaDepth dataset~\cite{li_megadepth_2018}, indoor environments from the ARKitScenes~\cite{dehghan2021arkitscenes} dataset and synthetic data generated using the Habitat-Sim simulator~\cite{habitat-sim}. 
In particular, we find that CroCo pretraining is a key ingredient to the success of our approach.
On the Aachen Day-Night~\cite{aachen} and Cambridge-Landmarks~\cite{KendallICCV15PoseNetCameraRelocalization} benchmarks, \ours{} outperforms current scene-specific and dataset-specific \SCR{} methods, while being competitive with state-of-the-art structure-based methods~\cite{kapture}.

\section{Related work}
\label{sec:related}

\PAR{Scene-specific coordinates regression.}
Several methods have been proposed to estimate dense 3D coordinates for a query image in a scene known at training time.
Early approaches~\cite{shotton_scene_2013, random_forest_2014, random_forest_2015} used regression forest models to predict the correspondence of a pixel in a RGB-D frame to its 3D world coordinate. 
More recent works~\cite{dsac, brachmann_learning_2018, esac, dsacstar, li2020hscnet, yang_sanet_2019, tang_learning_2021, kfnet, sc_wls_eccv2022,dong2022visual, huang2021vs} have replaced regression forests with CNN-based models that only require an RGB image. 
For example, Brachmann~\etal{}~\cite{dsac, brachmann_learning_2018, esac, dsacstar} train neural networks for this task and combine them with a differentiable RANSAC strategy for camera relocalization. 
Dong~\etal{}~\cite{dong2022visual} and Li~\etal{}~\cite{li2020hscnet} later introduce region classification into their pipelines for effective scene memorization. 
Huang~\etal{}~\cite{huang2021vs} propose to add a segmentation branch to obtain segmentation on scene-specific landmarks, which can then be associated with 3D landmarks in the scene to estimate camera pose. 
These methods are designed to memorize specific scenes,
making them hard to scale and impractical in many scenarios where the test scene is unknown at training time.
In contrast, our method can adjust at test time to any environment for which a database of images is available, by relying on external image retrieval techniques.

\PAR{Scene-agnostic coordinates regression with dense database 3D points.}
More related to our work are the scene-agnostic methods of~\cite{yang_sanet_2019,tang_learning_2021}. 
They regress dense scene coordinates given some reference views for which dense coordinates are already available. 
Their methods are also limited to small scenes with unit-normalized world coordinates.
In contrast, our approach only requires sparse annotations and imposes no restriction on coordinate range, making it better suited to large-scale environments.

\PAR{Image-based localization}
consists in estimating 6-DoF camera pose from a query image.
Different approaches can be used towards that goal, and \SCR{} is one of them.
Recently, learning-based methods in which the pose of a query image is directly regressed with a neural network have been proposed~\cite{KendallICCV15PoseNetCameraRelocalization,KendallCVPR17GeometricLossCameraPoseRegression,BrahmbhattCVPR18GeometryAwareLocalization,WangAAAI20AtLocAttentionGuidedCameraLocalization,lstmposenet,mspn}.
By training the network with database images and their known ground-truth poses as training set, they learn and memorize the relationship between RGB images and associated camera pose. 
These direct approaches however need to be trained specifically for each scene.
This issue was somehow solved by relative pose regression models~\cite{balntas2018relocnet,DingICCV19CamNetRetrievalForReLocalization,zhou2020essnet,arnold_map-free_2022}, which train a neural network to predict the relative pose between the query image and similar database image found by image retrieval. 
However, their performance tends to be inferior to structure-based methods~\cite{SchonbergerCVPR16StructureFromMotionRevisited,SnavelyIJCV08ModelingTheWorldFromInternetPhotoCollections,HeinlyCVPR15ReconstructingTheWorldSixDays,kapture}. %
Structure-based visual localization frameworks use sparse feature matching to estimate the pose of a query image relative to a 3D map constructed from database images using SfM techniques, such as those employed in~\cite{SchonbergerCVPR16StructureFromMotionRevisited}.
This involves extracting 2D features from images using interest point detectors and descriptors~\cite{sift, orb, superpoint, d2net, r2d2, disk, li2022decoupling_posfeat, aslfeat, sosnet, caps, lfnet, lift, geodesc}, and establishing 2D-3D correspondences. 
A %
PnP problem is solved using variants of RANSAC~\cite{fischlerrandom1981}, which then returns the estimated pose. 
However, the structure-based methods have to store not only 3D points but also keypoints descriptors, and maintaining the overall localization pipeline is complex.
Our approach in contrast does not require to store keypoints descriptors, is arguably simpler, %
and can use highly compressed database representations, thus reducing the storage requirement. 

\PAR{Database compression for visual localization.}
Compressing the database while maintaining localization accuracy is important for scalable localization systems. 
For structured-based methods, most techniques rely on selecting a compact but expressive subset of the 3D points.
K-cover method and its follow-up works~\cite{k_cover,prob_k_cover,weighted_k_cover,hybrid_scene_compression} reduce the number of 3D scene points, maintaining even spatial distribution and high visibility. 
Some other methods~\cite{covex_qp, mixed_qp, scale_qp, yang2022scenesqueezer} formulate the problem with quadratic programming (QP) to optimize good spatial coverage and visual distinctiveness.
Another approach for compression is feature quantization: descriptors associated to 3D points can be compressed into binary representation~\cite{cheng2019cascaded} or using quantized vocabularies~\cite{vocab_quan}.
In the field of SCR, the recent NeuMap~\cite{neumap} approach leverages a latent code per voxel and applies code-pruning to remove redundant codes.

\setlength{\abovedisplayskip}{5pt}
\setlength{\belowdisplayskip}{5pt}

\section{The \ours model}
\label{sec:method}

After describing our scene-agnostic coordinate regression model (Section~\ref{sub:model}), we then detail our robust coordinate encoding and associated training loss (Section~\ref{sub:loss}).
We finally present the application to visual localization (Section~\ref{sub:visloc}) %
and training details (Section~\ref{sub:training}).

\begin{figure*}[t]
\centering
\resizebox{\linewidth}{!}{
\begin{SCRfigtabular}
\SCRfigline{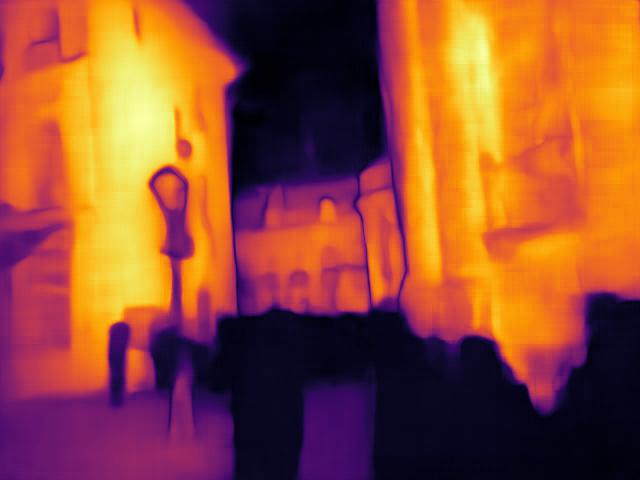} \\
\SCRfigline{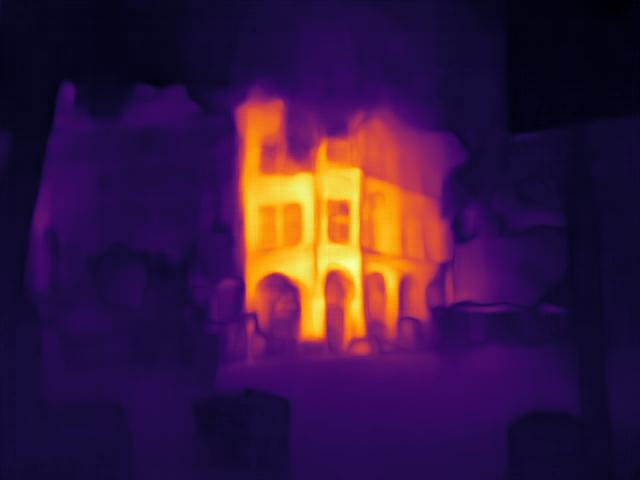} \\
\SCRfigline{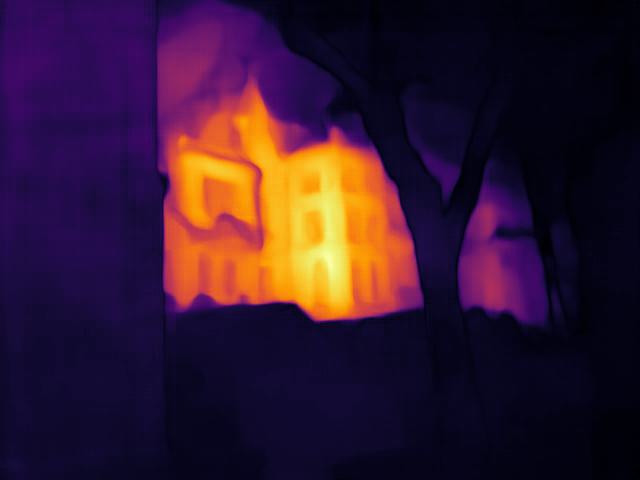} \\
\end{SCRfigtabular}} \\[-0.3cm]
\caption{\label{fig:regression_results}\textbf{Regression examples on Aachen-Day.}
Our model predicts a dense 3D coordinates point map and a confidence map (\nth{5} and \nth{6} columns) for a given query image (\nth{4} column) using reference images retrieved from a \SfM{} database (\nth{1}, \nth{2}, \nth{3} columns). Only the first 3 reference images (out of $K=8$) are depicted.
3D coordinates and confidence are colorized for visualization purposes, and areas with a confidence below $\tau=\exp(0)$ are not displayed.
Best viewed in color.
}
\vspace{-0.2cm}

\end{figure*}

\subsection{Model architecture}
\label{sub:model}

Our model takes as input a query image $\I_q$ and a mapped database images $\I_b$ for which sparse 2D-3D annotations are available, denoted as $\V = \{(\p_j, \pv_j)\}$ where $\pv_j\in\mathbb{R}^3$ is a 3D point expressed in a world coordinate system visible at pixel $\p_j$. 
It then predicts a 3D coordinate point for every pixel in the query image.
In the more realistic case where multiple database images are relevant to the query, we perform independent predictions between the query and each database image with the model described below, and fuse predictions afterward (see Section~\ref{sub:visloc}).

\PAR{Overview.}
Figure~\ref{fig:crocordinate} shows an overview of the model architecture. 
First, the query image is encoded into a set of token features with a Vision Transformer~\cite{vit} (ViT) encoder. 
The same encoder is used to encode the database image, but this time the resulting database features are augmented with geo-spatial information from the sparse 2D-3D annotations.
This is achieved using a transformer decoder %
referred to as \emph{3D Mixer} in the following.
The next step consists in transferring geo-spatial information from the augmented database features to the query features using a transformer decoder.
Finally, a prediction head outputs dense 3D coordinates for each pixel of the query image, see Figure~\ref{fig:regression_results}. %
We now detail each module: the image encoder, the 3D mixer, the decoder and the prediction head.

\PAR{Image encoder.}
We use a vision transformer~\cite{vit} to encode the query and database images.
In more details, each image is divided into non-overlapping patches, and a linear projection encodes them into patch features. 
A series of transformer blocks is then applied on these features: each block consists of multi-head self-attention and an MLP. 
In practice, we use a ViT-Base model, \ie, $16{\times}16$ patches with $768$-dimensional features, $12$ heads and $12$ blocks.
Following~\cite{deepmatcher,crocostereo}, we use RoPE~\cite{rope} relative position embeddings.
As a result of the ViT encoding, we obtain sets of token features denoted $\R_q$ for the query and $\R_b$ for the database image respectively.

\PAR{3D mixer.}
We then augment the database tokens $\R_b$ with geo-spatial information encoded by the sparse 2D-3D correspondence set $\V$, yielding the augmented tokens $\Raug_b = \text{3Dmixer}\left( \R_b, \V \right)$.
To that aim, we encode the 3D coordinates using a cosine point encoding $\phi$ before feeding them to an MLP (see next Section~\ref{sub:loss} for details on $\phi$).
We then use a series of transformer decoder blocks, where each block consists of a self-attention between image tokens, %
a cross-attention that shares information from all point tokens with these image tokens, and an MLP.
We find that alternating between image-level and point-level decoder blocks improves the performance, and we refer to Appendix~\ref{sec:3dmixer} for more details and ablative studies on the 3D mixer architecture.

\PAR{Decoder.}
The next step is to transfer information from the database, \ie, from ${\Raug_b}$, into the query features $\R_q$. 
We again rely on the cross-attention mechanism of a generic transformer decoder, \ie, a series of blocks, each composed of self-attention between the token features, cross-attention with %
the database tokens $\Raug_b$
and an MLP, yielding augmented query features $\Raug_q$.

\PAR{Prediction head.}
We finally reshape $\Raug_q$ from the last transformer decoder block into a dense feature map and apply a convolutional head.
Specifically, we first linearly project the features to 1024 dimensions, then apply a sequence of 6 ConvNeXt blocks~\cite{liu2022convnet}, with a PixelShuffle~\cite{pixelshuffle_superresol} operation every two blocks to increase the resolution while halving the channel dimension.
For a $224\times 224$ input image, we get a $14^2 \times 1024$ token map after the initial projection, which is gradually expanded to $28^2 \times 512$, $56^2 \times 256$ and finally $224^2 \times d$, $d$ being the output dimension.

\subsection{Generalization and training loss}
\label{sub:loss}

\newcommand{\hv}[0]{\hat{\textbf{v}}}

\PAR{Output space.}
A naive approach consists in setting $d=3$, \ie, trying to directly regress dense 3D points $\{\hv\}\in\mathbb{R}^3$ from the regression head.
This is possible and could be trained with a standard $\ell_1$ or $\ell_2$ regression loss, but is subject to a major limitation.
At test time, the network is typically unable to regress coordinates outside the range seen during training.
Thus, except for small scenes, it cannot generalize to new datasets (see Section~\ref{sub:ablate}).

Instead,  we propose to regress
a higher-dimensional 3D point encoding $\phi(\pv) \in (\mathbb{S}^1)^{d/2} \subset [-1,1]^d$, with $d \gg 3$.
We design $\phi$ with several desirable properties holding for any given $\pv\in\mathbb{R}^3$:
(i) $\phi$ is an injective mapping, with an inverse projection $\phi^{-1}$ such that $\phi^{-1}(\phi(\pv))=\pv$; 
(ii) the input space of $\phi^{-1}$ is %
the unit-circle product  $(\mathbb{S}^1)^{d/2} \subset [-1,1]^d$, whose high dimension enables error-correcting mechanisms in $\phi^{-1}$.
Thanks to these properties, our method can handle any coordinate at test time.

\PAR{Point encoding.}
Assuming uncorrelated $x$, $y$ and $z$ coordinates, we can decompose $\phi(\pv) = [\psi(x),\psi(y),\psi(z)]$ and define $\psi(x)$ as:
\begin{equation}
    \psi(x) = \left[\cos(f_1 x), \sin(f_1 x), \cos(f_2 x), \sin(f_2 x), \ldots \right] 
\end{equation}
where the $f_i$'s are frequencies defined as $f_i=f_0 \gamma^{i-1}$, $i \in \{1, \ldots, d/6\}$, with $f_0 >0$ and $\gamma > 1$.
In practice, we set $f_0$ and $\gamma$ such that the periods of the lowest and highest frequencies $f_1$ and $f_{d/6}$ approximately correspond to the maximum scale of a query scene (\eg 300 meters) and the desired spatial resolution (\eg 0.5 meter).
The encoding dimension $d$ then becomes a parameter that controls the level of redundancy.
$d$ must be carefully chosen, as too small encodings are not noise-resistant enough, while too large encodings may demand too much capacity for the decoder.
The inverse mapping $\psi^{-1}$ efficiently solves a least-square problem of the form %
$\psi^{-1}(y) = \text{argmin}_x \left\Vert y-\psi(x)\right\Vert^2$, see Appendix~\ref{sup:inverse} and~\cite{cosine}.

\PAR{Regression loss.}
As for the naive regression case, we apply a standard $\ell_1$ regression loss to train the network:
\begin{equation}
    \mathcal{L}_{\text{reg}}(\pv, \y) = \left\vert \phi(\pv) - \y \right\vert,
    \label{eq:loss_reg}
\end{equation}
where $\y \in \mathbb{R}^d$ is the network output and $\pv$ is the corresponding ground-truth 3D point.
We further exploit the relation between pairs of adjacent %
components of $\phi(\pv)$, based on the equality $\cos^2(f_ix) + \sin^2(f_i x) = 1$. 
Before applying the $\mathcal{L}_{\text{reg}}$ loss, we thus $\ell_2$-normalize each pairs of consecutive %
components of $\y$.
We empirically find that this helps the training significantly.

\PAR{Pixelwise confidence.}
Regressing coordinates is inevitably harder, or even impossible, for some parts of the query image
such as the sky or objects not visible in database images. 
We therefore jointly predict a per-pixel confidence $\tau > 0$ that modulates the regression loss~\eqref{eq:loss_reg}, following~\cite{kendall2018multi}:
\begin{equation}
    \mathcal{L}_{\text{SCR}}({\pv, \y}, \tau) = %
    \tau \mathcal{L}_{\text{reg}}({\pv,\y}) - \log \tau.
    \label{eq:loss}
\end{equation}
$\tau$ can be interpreted as the confidence of the prediction: if $\tau$ is low for a given pixel, the corresponding $\mathcal{L}_{\text{reg}}$ loss at this location will be down-weighted. 
The second term of the loss incites the model to avoid being under-confident.
The estimated confidence can also serve to fuse predictions from multiple database images, as well as for the PnP pose estimation step, see Section~\ref{sub:visloc}.

\subsection{Application to visual localization}
\label{sub:visloc}

We now present how our model can be applied to predict the camera pose of a given query image from a small set of relevant database images with sparse 2D-3D point correspondences.
An overview of our visual localization pipeline is shown in Figure~\ref{fig:overview}.

\PAR{Image retrieval.}
Given a query image, we first follow the same retrieval step than for standard feature-matching-based localization approaches~\cite{kapture,hfnet,SattlerPriorityLoc2017}. 
Namely, we utilize off-the-shelf image retrieval methods such as HOW~\cite{how}, AP-GeM~\cite{apgem} or FIRe~\cite{fire} to obtain a shortlist of $K$ relevant database images for a given query image.

\PAR{Sparse 2D-3D annotations.}
Our model takes as input sparse 2D-3D correspondences for each database image.
To get them, we randomly subsample 2D points from the dense RGB-D data and reproject them in 3D using the known camera poses, when available. If not, we rely on standard Structure-from-Motion pipelines~\cite{SchonbergerCVPR16StructureFromMotionRevisited} during which
2D keypoint matches between images are used to recover the corresponding 3D point locations and the camera poses. 
This process directly yields a set of 2D-3D correspondences for each database image.
In practice, we use the output of COLMAP~\cite{SchonbergerCVPR16StructureFromMotionRevisited} with SIFT~\cite{sift} keypoints.

\PAR{Multi-image fusion strategy.}
To mitigate the potential presence of outliers returned by the image retrieval module, we fuse the predictions from the top-$K$ relevant database images.
We first compute the augmented database features $\Raug_b$ for each image $\I_b$ separately, with $b=1\ldots K$.
We then feed each ($\R_q$, $\Raug_b$) pair to the decoder, gathering each time the dense coordinate and confidence output maps. %
The final aggregation is then simply done pixelwise. 
We fuse all results by keeping, for each pixel $i$, the most confident prediction %
according to the estimated confidence $\{\tau_b^i\}_{b=1\ldots K}$. %

\PAR{Predicting camera poses.}
The output of our model is a dense 3D coordinate map and corresponding confidence map, see Figure~\ref{fig:crocordinate}.
To perform visual localization,
we first filter out all unconfident predictions, \ie, points for which the confidence is inferior to the median confidence.
We then use an off-the-shelf PnP solver to obtain the predicted camera pose.
Specifically, we rely on SQ-PnP~\cite{sqpnp} with 4096 2D-3D correspondences sampled randomly, 10,000 iterations and a reprojection error threshold of 5 pixels.

\PAR{Database compression.}
Since spatially-augmented database features $\Raug_b$ do not depend on the query image (see Figure~\ref{fig:crocordinate}), they can thus be computed offline once and stored.
Raw representations require a few megabytes (MB) of storage per database image, similar to standard feature-based localization methods.
We find however that they can be significantly compressed with negligible loss of performance.
Namely, we employ Product Quantization (PQ)~\cite{jegou_pq_2011}, which is a simple and effective technique consisting of splitting vectors into multiple sub-vectors and vector-quantizing~\cite{vector_quantization} them into byte codes (see Appendix~\ref{sup:compression} for more details).
Note that all the compression parameters (\eg codebooks) are scene-agnostic as well, \ie, trained once and for all.

\subsection{Training details}
\label{sub:training}

We initialize the weights of the encoder and the decoder with CroCo v2 pretraining~\cite{crocostereo}, which we find crucial for the success of our approach.
We train our model on $512{\times}384$ images, but perform a first training stage with $224{\times}224$ images while freezing the encoder, \ie, fine-tuning only the 3D mixer and the decoder for 100 epochs with a fixed learning rate of $10^{-4}$  to reduce overall training costs. Training is then performed at higher resolution for 40 epochs with a cosine decay learning rate schedule.

\PAR{Data.}
We train our model on datasets that cover various scenarios for robustness: MegaDepth~\cite{li_megadepth_2018} contains SfM reconstruction of 275 (mainly) outdoor scenes, ARKitScenes~\cite{dehghan2021arkitscenes} consists of indoor house scenes, and Habitat of synthetic indoor scenes derived from  HM3D~\cite{ramakrishnan2021hm3d}, ScanNet~\cite{dai2017scannet}, Replica~\cite{replica19arxiv} and ReplicaCAD~\cite{szot2021habitat} rendered using Habitat-Sim~\cite{habitat-sim}.
These three datasets provide %
dense depth estimates and camera poses, thus allowing to train our model in a fully-supervised manner. 
We use 100K query from each dataset (300K in total). %
For each query, we use FIRe~\cite{fire} to retrieve beforehand a shortlist of $K$ similar images.

\PAR{Augmentation.}
We 
apply standard random crop and color jitter during training.
For robustness to possible triangulation noise, we augment 5\% of the sparse 3D points with simulated depth noise.
We also apply random geometric 3D transformation to scene coordinates for better  generalization. %
Namely, we apply random 3D rotation followed by random scaling in the range $[1/2,2]$ and random translation in $[-1000m,1000m]^3$.

\section{Experiments}

After describing the test datasets (Section~\ref{sub:datasets}), we present ablations in Section~\ref{sub:ablate} and provide visualizations of the attention in Section~\ref{sub:cross_attn}.
We then compare our approach to the state of the art in visual localization without (Section~\ref{sub:sota}) and with compression (Section~\ref{sub:compression}), and finally evaluate the accuracy of the regressed coordinates (Section~\ref{sub:coordinates_regression}).

\subsection{Datasets and metrics}
\label{sub:datasets}

\PAR{Cambridge-Landmarks}~\cite{KendallICCV15PoseNetCameraRelocalization}
consists of 6 outdoor scenes with RGB images from videos and small-scale landmarks. 

\PAR{7 Scenes}~\cite{ShottonCVPR13SceneCoordinateRegression}
consists of 7 indoor scenes with RGB-D images from videos. Each scene has a limited size, and the images contain repeating structures, motion blur, and texture-less surfaces.%
We do not use the depth data of the query image during inference.

\PAR{Aachen Day-Night v1.1}~\cite{aachen,ZhangIJCV20ReferencePoseGenerationVisLoc}
contains 6,697 database images captured at day time, and 1015 query images including 824 taken during daytime (Aachen-Day) and 191 during nighttime (Aachen-Night). 

\PAR{Metrics.} For Cambridge and 7-Scenes, we report the median translation error. For Aachen, we report the percentage of successfully localized images within three thresholds: (0.25m, 2\textdegree), (0.5m, 5\textdegree) and (5m, 10\textdegree).

\begin{table}[]
    \centering
    \resizebox{\linewidth}{!}{
    \begin{tabular}{c|c|ccc}
    \toprule 
    Point encoding & Aug & Camb.~$\downarrow$ & 7scenes~$\downarrow$ & Aachen-Night~$\uparrow$\\
    \midrule 
    $(x,y,z)\in\mathbb{R}^{3}$ &   & 1.69 & \textbf{0.11} & 0.0 / 0.0 / 0.0 \\
    $(x,y,z)\in\mathbb{R}^{3}$ &  \checkmark & 14.43 & 2.89  & 0.0 / 2.1 / 44.5 \\
    $\phi(\cdot)\in[-1,1]^{24}$ & \checkmark & 0.47 & \textbf{0.11} & 22.0 / 46.6 / 89.5 \\
    \hl $\phi(\cdot)\in[-1,1]^{36}$ & \hl \checkmark & \hl \textbf{0.43} & \hl \textbf{0.11 } & \hl 22.0 / \textbf{47.1} / \textbf{90.6 }\\
    $\phi(\cdot)\in[-1,1]^{48}$ & \checkmark & 0.55  & \textbf{0.11 } & \textbf{23.6} / 40.8 / 87.4 \\
    \bottomrule
    \end{tabular}
    } \\[-0.25cm]
    \caption{\textbf{Ablation on 3D point encoding.} Aug=Augmentation.
    \vspace{-0.2cm}
        }
    \label{tab:ab_encoding}
\end{table}

\begin{table}[]
    \centering
    \resizebox{0.9\linewidth}{!}{
    \begin{tabular}{cc|ccc}
    \toprule
    Pretraining & Frozen & Camb.~$\downarrow$ & 7scenes~$\downarrow$ & Aachen-Night~$\uparrow$\\
    \midrule 
    - & -       & 1.14  & 0.19  & 5.2 / 20.4 / 66.0\\
    CroCo v2 & -       & 0.54  & 0.14  & 18.3 / 37.7 / 85.3\\
    \hl CroCo v2 & \hl Encoder & \hl \textbf{0.43}  & \hl \textbf{0.11} & \hl \textbf{22.0} / \textbf{47.1} / \textbf{90.6} \\
    \bottomrule
    \end{tabular}
    } \\[-0.25cm]
    \caption{\textbf{Ablation on pretraining and encoder freezing.}
    \vspace{-0.3cm}
        }
    \label{tab:ab_pretraining}
\end{table}

\subsection{Ablative study}
\label{sub:ablate}

We now ablate the main design, architectural and and training choices of our approach.
We perform all ablations using a lower image resolution of $224{\times}224$ with a single retrieved image ($K=1$). 
For each ablation table, we put a gray background color on the row with default settings.

\PAR{Validation sets and metrics.}
We report the visual localization performance on a selected subset of 5 diverse and relatively challenging datasets: 7scenes-stairs, 7scenes-pumpkin, Cambridge-GreatCourt, Cambridge-OldHospital and Aachen-Night. 
For 7scenes and Cambridge-Landmarks, we report the averaged median translation error, while for Aachen-Night we report the localization accuracy for the 3 standard thresholds.

\PAR{Robust coordinate encoding.}
We first study in Table~\ref{tab:ab_encoding} the impact of different point encoding schemes.
Notably, we observe that direct coordinate regression is only successful when the train and test output distributions are aligned.
This is the case for 7-scenes, or Cambridge to a lesser extent, as they are small and well-centered around the origin. 
For larger scenes with unconstrained coordinates (like Aachen), direct regression utterly fails.
One way to mitigate this issue is to augment 3D coordinates at training time, \eg using random translations (see Section~\ref{sub:training}).
Augmentations somehow improve the situation for Aachen-Night, but the performance overall strongly degrades for Cambridge and 7scenes.
In contrast, the cosine-based encoding $\phi$ proposed in Section~\ref{sub:loss} effectively deals with indoor and outdoor scenes in any coordinate ranges.
We find optimal to use $6$ frequencies, yielding $d=36$-dimensional outputs.

\PAR{Impact of CroCo pretraining.}
Table~\ref{tab:ab_pretraining} shows that pretraining the ViT encoder and decoder with CroCo v2~\cite{crocostereo} self-supervised objective is key to the success of our approach.
Without CroCo pretraining, the performance significantly drops, which is explained by the fact that CroCo essentially learns to compare and implicitly match images, which is empirically verified in Section~\ref{sub:cross_attn}.
We hypothesize that CroCo pretraining also ensures generalization, since the pretraining set (7M pairs) is much larger than our training dataset.
Another illustration of this benefit is that the performance further improves when we \emph{freeze} the ViT encoder during this training step, meaning that pretraining with CroCo effectively learns image representations already fit for our coordinate regression task.

\begin{table}[]
    \centering
    \resizebox{\linewidth}{!}{
    \begin{tabular}{cc|ccc}
    \toprule
    Regression head & Channels & Camb.~$\downarrow$ & 7scenes~$\downarrow$ & Aachen-Night~$\uparrow$\\
    \midrule 
    Linear            & $x,y,z,\tau$  & 0.94 & 0.12 &	11.0 / 31.9 / 84.3 \\
    ConvNeXt & $xyz\tau$     & 0.64 & \textbf{0.11} &	19.9 / 39.8 / 88.0 \\
    ConvNeXt & $xyz,\tau$  & 0.61 & \textbf{0.11} &	15.7 / 41.4 / 87.4 \\
    \hl ConvNeXt & \hl$x,y,z,\tau$  & \hl\textbf{0.43} & \hl\textbf{0.11} &	\hl\textbf{22.0} / \textbf{47.1} / \textbf{90.6} \\
    \bottomrule
    \end{tabular}
    }\\[-0.25cm]
    \caption{\textbf{Ablation on regression head.}}
    \vspace{-0.2cm}
    \label{tab:ab_head}
\end{table}

\begin{table}[]
    \centering
    \newcommand{\blocks}[0]{b}
    \resizebox{\linewidth}{!}{
    \begin{tabular}{cc|ccc}
    \toprule
    Train res. & Test res. & Camb.~$\downarrow$ & 7scenes~$\downarrow$ & Aachen-Night~$\uparrow$\\
    \midrule 
    \hl $224{\times}224$   & \hl $224{\times}224$  & \hl 0.43 & \hl 0.11 &	\hl 22.0 / 47.1 / 90.6 \\
    $512{\times}384$   & $512{\times}384$  & 0.21 & 0.10 &	39.3 / 63.4 / \textbf{94.8}  \\
    $512{\times}384$   & $640{\times}480$  & \textbf{0.20} & \textbf{0.07} & \textbf{45.5} / 68.6 / \textbf{94.8}  \\
    $512{\times}384$   & $768{\times}512$  & 0.24 & \textbf{0.07} & {\bf 45.5} / \textbf{70.2} / 93.7 \\
    \bottomrule
    \end{tabular}
    } \\[-0.25cm]
    \caption{\textbf{Impact of training and test image resolution.}}
    \label{tab:img_res}

    \vspace{-0.3cm}
    
\end{table}

\PAR{Separate heads.}
We experiment with different architectures for the regression head, this time aiming at exploiting priors of the output space.
Recall that for each pixel, we ultimately predict 4 values: 3 spatial components ($x$, $y$ and $z$) and a confidence $\tau$.
A priori, these four components have no reason to be correlated. 
In fact, predicting them jointly could turn detrimental if there is a risk for the network to learn false correlations.
Therefore, we compare:  (i) as a baseline, a simple linear head, which is the same as 4 independent linear heads (one per component); (ii) regressing the 4 components jointly using the same head; (iii) regressing the spatial and confidence components separately;
(iv) regressing all 4 components separately, in which case we still use the same prediction head with shared weight for all spatial $x$, $y$ and $z$ components after an independent linear projection.
From Table~\ref{tab:ab_head}, option (iv) clearly yields the best performance, while the linear heads is the worst option.

\PAR{Image resolution
} 
can have a strong impact on the test performance.
Table~\ref{tab:img_res} shows that test performance generally increases as a function of image resolution.
Interestingly, the model is able to generalize to higher resolution at test time, as training in $512\times 384$ and testing on higher resolutions consistently yields better results.
In the following, we always test on $640\times 480$ images.

\subsection{Visualization of internal attention }
\label{sub:cross_attn}

\begin{figure}
    \centering
    \resizebox{\linewidth}{!}{
    \hspace{-25mm}
    \raisebox{-0.5\height}{\includegraphics{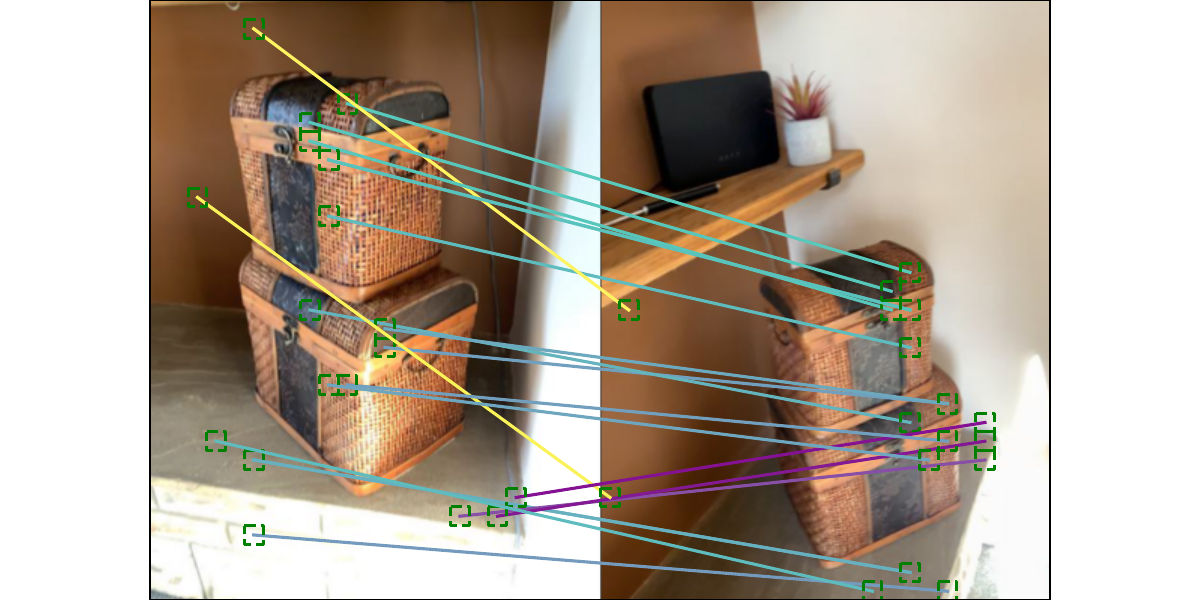}}
    \hspace{-20mm}
    \raisebox{-0.5\height}{\includegraphics[height=135mm]{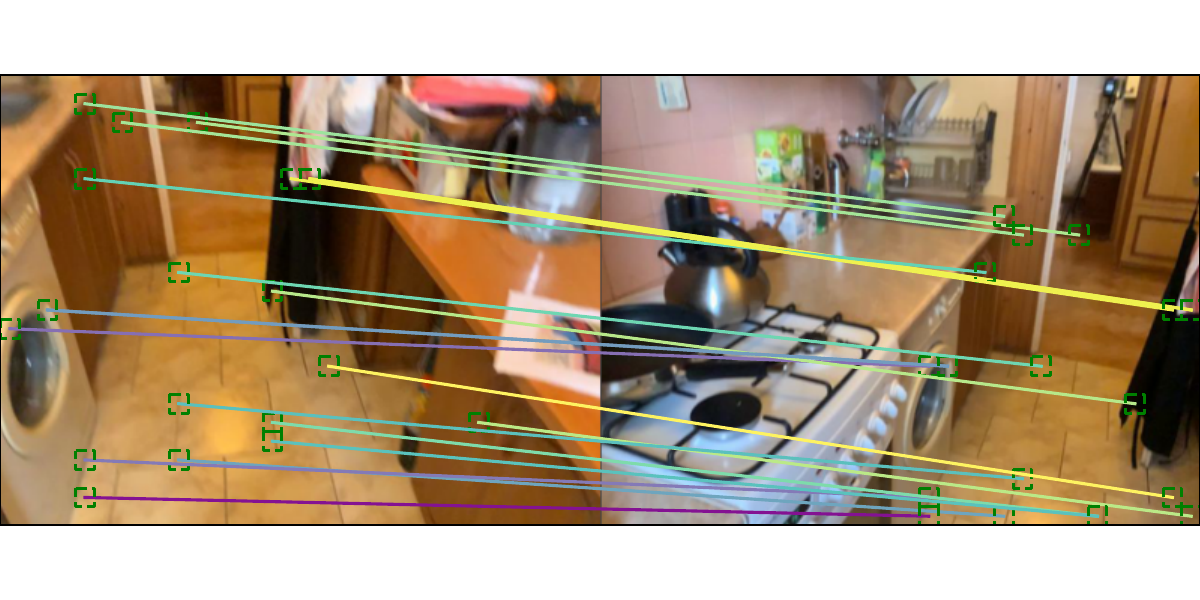}}
    }
    \vspace{-6mm}
    \caption{\textbf{Two example visualizations of the cross attention between query and reference (left and right image, resp.) images in the decoder.}
    We plot the top-20 cross-attention scores as red lines between 16x16 image patches of their corresponding tokens. }
    \vspace{-0.35cm}
    \label{fig:cross_attn}
\end{figure}

To better understand how the network is able to perform the coordinate regression task, we visualize in Figure~\ref{fig:cross_attn} the highest cross-attention scores in the decoder, displayed as patch correspondences between corresponding tokens.
Interestingly, we observe that the decoder implicitly performs image matching under the hood.
In a sense, this is expected since to solve the task, the model has to essentially perform a matching-guided %
interpolation/extrapolation
of the known reference coordinates to the query image.
Note that it learns to implicitly perform matching without any explicit supervision for this task (\ie, only from the regression signal).

\subsection{Visual localization benchmarking}
\label{sub:sota}

\begin{figure*}[]
    \centering
    \begin{subfigure}[b]{0.74\linewidth}
        \resizebox{\linewidth}{!}{
        \small
        \begin{tabular}{c@{~~}l|ccccccc}
        \toprule 
         &  & Chess~$\downarrow$ & Fire~$\downarrow$ & Heads~$\downarrow$ & Office~$\downarrow$ & Pumpkin~$\downarrow$ & Kitchen~$\downarrow$ & Stairs~$\downarrow$\\
        \midrule
        \multirow{2}{*}{\begin{turn}{90}
        Kpts
        \end{turn}} & Active search \cite{SattlerPriorityLoc2017} & 0.04, 1.96 & \uline{0.03}, 1.53 & \uline{0.02}, 1.45 & 0.09, 3.61 & 0.08, 3.10 & 0.07, 3.37 & \textbf{0.03}, 2.22\\
         & HLoc \cite{hfnet} & \textbf{0.02}, 0.79 & \textbf{0.02}, \uline{0.87} & \uline{0.02}, 0.92 & \textbf{0.03}, 0.91 & \uline{0.05}, 1.12 & \uline{0.04}, 1.25 & 0.06, 1.62\\
        \midrule 
        \multirow{11}{*}{\rotatebox[origin=c]{90}{Learning-based}}
        & RelocNet~\cite{balntas2018relocnet} & 0.12, 4.14 & 0.26, 10.4 & 0.14, 10.5 & 0.18, 5.32 & 0.26, 4.17 & 0.23, 5.08 & 0.28, 7.53 \\
        & CamNet~\cite{DingICCV19CamNetRetrievalForReLocalization} & 0.04, 1.73 & \uline{0.03}, 1.74 & 0.05, 1.98 & \uline{0.04}, 1.62 & \textbf{0.04}, 1.64 & \uline{0.04}, 1.63 & \uline{0.04}, 1.51\\
        & DSAC++ \cite{brachmann_learning_2018} & \textbf{0.02}, \textbf{0.5} & \textbf{0.02}, 0.9 & \textbf{0.01}, \textbf{0.8} & \textbf{0.03}, \uline{0.7} & \textbf{0.04}, 1.1 & \uline{0.04}, \textbf{1.1} & 0.09, 2.6\\
        & KFNet \cite{kfnet} & \textbf{0.02}, \uline{0.65} & \textbf{0.02}, 0.9 & \textbf{0.01}, \uline{0.82} & \textbf{0.03}, \textbf{0.69} & \textbf{0.04}, \uline{1.02} & \uline{0.04}, 1.16 & \textbf{0.03}, \uline{0.94}\\
        & HSCNet \cite{li2020hscnet} & \textbf{0.02}, 0.7 & \textbf{0.02}, 0.9 & \textbf{0.01}, 0.9 & \textbf{0.03}, 0.8 & \textbf{0.04}, \textbf{1.0} & \uline{0.04}, 1.2 & \textbf{0.03}, \textbf{0.8}\\
         & SANet \cite{yang_sanet_2019} & \uline{0.03}, 0.88 & \uline{0.03}, 1.12 & \uline{0.02}, 1.48 & \textbf{0.03}, 1.00 & \textbf{0.04}, 1.21 & \uline{0.04}, 1.40 & 0.16, 4.59\\
         & DSM \cite{tang_learning_2021} & \textbf{0.02}, 0.68 & \textbf{0.02}, \textbf{0.80} & \textbf{0.01}, \textbf{0.8} & \textbf{0.03}, 0.78 & \textbf{0.04}, 1.11 & \textbf{0.03}, \uline{1.11} & \uline{0.04}, 1.16\\
         & SC-wLS \cite{sc_wls_eccv2022} & \uline{0.03}, 0.76 & 0.05, 1.09 & 0.03, 1.92 & 0.06, 0.86 & 0.08, 1.27 & 0.09, 1.43 & 0.12, 2.80\\
         & NeuMaps \cite{neumap} & \textbf{0.02}, 0.81 & \uline{0.03}, 1.11 & \uline{0.02}, 1.17 & \textbf{0.03}, 0.98 & \textbf{0.04}, 1.11 & \uline{0.04}, 1.33 & \uline{0.04}, 1.12\\
         & \textbf{\ours}, K=20 & \uline{0.03}, 0.94 & \uline{0.03}, 1.12 & \uline{0.02}, 1.08 & \uline{0.04}, 1.10 & \uline{0.05}, 1.38 & 0.05, 1.36 & 0.05, 1.44\\
         & \textbf{\ours-L}, K=20 & \uline{0.03}, 0.94 & \uline{0.03}, 1.03 & \uline{0.02}, 1.16 & \textbf{0.03}, 1.06 & \uline{0.05}, 1.41 & \uline{0.04}, 1.35 & 0.06, 1.62\\
        \bottomrule
        \end{tabular}
        }
    \end{subfigure}
    \hfill
    \begin{subfigure}[b]{0.24\linewidth}
        \centering
        \resizebox{\linewidth}{!}{ 
            \includegraphics{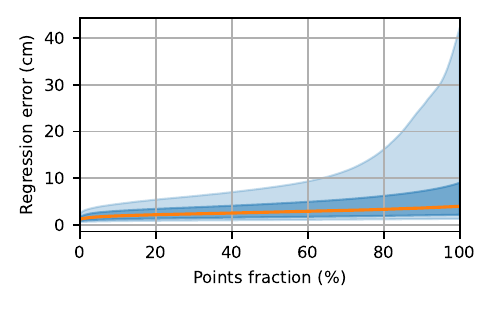} 
        }
        \vspace{0mm}
        \resizebox{\linewidth}{!}{ 
            \includegraphics{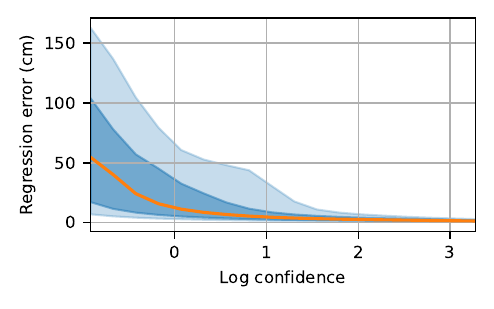}
        }
        \vspace{-30mm}
    \end{subfigure}
    \vspace{-0.28cm}
    \caption{
        \label{fig:7-scenes}
        \textbf{Evaluations on 7-Scenes.}
        \emph{Left: visual localization.} 
        Comparison with the state of the art in terms of median translation (m) and angular ($^\circ$) errors.
        \emph{Right: SCR.} %
        Distribution of coordinate prediction errors (first and last quartiles, deciles, and median) \wrt relative 
        (top plot where we keep the $x\%$ most confident predicted points for each image)
        and absolute confidence (bottom plot where we show statistics for all points with a confidence above a given threshold).
        Errors are typically below 10cm, and correlate well with the predicted confidence.
        \vspace{-0.5cm}
        } 
\end{figure*}

We compare our approach to the state of the art for visual localization on indoor (7-scenes) and outdoor datasets (Cambridge-Landmarks, Aachen-DayNight).
We compare to learning-based approaches as well as a few representative keypoint-based methods such as Active Search~\cite{SattlerPriorityLoc2017} and HLoc~\cite{hfnet}.
Results are presented in Table~\ref{tab:sota}, Table~\ref{tab:cambridge} and Figure \ref{fig:7-scenes},
with \emph{SACReg} and \emph{SACReg-L} denoting the proposed method using a ViT-Base or ViT-Large encoder backbone respectively.
On the indoor 7-Scenes dataset, our method obtains similar or slightly worse performance compared to other approaches, but overall still performs well with a median error of a few centimeters.
On outdoor datasets, the proposed methods strongly outperforms other learning-based methods, in particular other scene-specific or scene-agnostic coordinate regression approaches like~\cite{dsac,dsacstar,sc_wls_eccv2022,yang_sanet_2019,tang_learning_2021,neumap,esac}.
This is remarkable because, in contrast to any other learning-based approaches, \ours is directly applied to each test set without any finetuning.
In other words, our approach works out of the box on test data that were never seen during training.
Interestingly, it even reaches the performance of keypoints-based approaches such as Active Search~\cite{SattlerPriorityLoc2017} or HLoc~\cite{hfnet}. 

\begin{table}
    \centering
    \resizebox{\linewidth}{!}{
    \begin{tabular}{l@{~~}l|cc}
    \toprule
    &  & Aachen-Day~$\uparrow$ & Aachen-Night~$\uparrow$ \\
    \midrule
    \multirow{2}{*}{\begin{turn}{90}
    Kpts
    \end{turn}} & Active Search~\cite{SattlerPriorityLoc2017}         & 57.3 / 83.7 / 96.6 & 28.6 / 37.8 / 51.0 \\
    & HLoc~\cite{hfnet}       & \textbf{89.6} / \textbf{95.4} / 98.8 & \textbf{86.7} / \textbf{93.9} / \textbf{100} \\
    \midrule
    \multirow{6}{*}{\begin{turn}{90}
    Learning-based
    \end{turn}} & DSAC~\cite{dsac} & 0.4 / 2.4 / 34.0 & - \\
    & ESAC (50 experts)~\cite{esac} & 42.6 / 59.6 / 75.5 & - \\
    & HSCNet~\cite{li2020hscnet} & 65.5 / 77.3 / 88.8 & 22.4 / 38.8 / 54.1 \\
    & NeuMap~\cite{neumap} & 76.2 / 88.5 / 95.5 & 37.8 / 62.2 / 87.8 \\
    \cmidrule{2-4}
    & \textbf{\ours{}}, $K=20$ & \uline{85.3} / \uline{93.7} / \textbf{99.6} & \uline{64.9} / \uline{90.1} / \textbf{100.0} \\
    & \textbf{\ours{}-L}, $K=20$ & \uline{85.8} / \uline{95.0} / \textbf{99.6} & \uline{67.5} / \uline{90.6} / \textbf{100.0} \\ 
    \bottomrule
    \end{tabular}
    } \\[-0.25cm]
    \caption{\textbf{Comparison to the state of the art on Aachen.}%
    }
    \label{tab:sota}

    \vspace{-0.2cm}
    
\end{table}

\begin{table}[]
    \centering
    \resizebox{\linewidth}{!}{
    \begin{tabular}{c@{~~}l|ccccc}
    \toprule 
     &  & {\footnotesize ShopFacade$\downarrow$}  & {\footnotesize OldHospital$\downarrow$} & {\footnotesize College$\downarrow$} & {\footnotesize Church$\downarrow$} & {\footnotesize Court$\downarrow$}\\
    \midrule 
    \multirow{2}{*}{\begin{turn}{90}
    Kpts
    \end{turn}} & Active search~\cite{SattlerPriorityLoc2017} & 0.12, 1.12 & 0.52, 1.12 & 0.57, 0.70 & 0.22, 0.62 & 1.20, 0.60\\
     & HLoc~\cite{hfnet} & \textbf{0.04,} \textbf{0.20} & \uline{0.15}, 0.3 & \uline{0.12}, 0.20 & \uline{0.07}, \uline{0.21} & \uline{0.11}, 0.16\\
    \midrule 
    \multirow{10}{*}{\begin{turn}{90}
    Learning-based
    \end{turn}} & DSAC++~\cite{brachmann_learning_2018} & 0.06, 0.3 & 0.20, 0.3 & 0.18, 0.3 & 0.13, 0.4 & 0.20, 0.4\\
     & DSAC*~\cite{dsacstar}  & \uline{0.05}, 0.3 & 0.21, 0.4 & 0.15, 0.3 & 0.13, 0.4 & 0.49, 0.3\\     
     & KFNet~\cite{kfnet}  & \uline{0.05}, 0.35 & 0.18, \uline{0.28} & 0.16, 0.27 & 0.12, 0.35 & 0.42, 0.21\\     
     & HSCNet~\cite{li2020hscnet}  & 0.06, 0.3 & 0.19, 0.3 & 0.18, 0.3 & 0.09, 0.3 & 0.28, 0.2\\
     & SANet~\cite{yang_sanet_2019}  & 0.1, 0.47 & 0.32, 0.53 & 0.32, 0.54 & 0.16, 0.57 & 3.28, 1.95\\     
     & DSM~\cite{tang_learning_2021} & 0.06, 0.3 & 0.23, 0.38 & 0.19, 0.35 & 0.11, 0.34 & 0.19, 0.43\\
     & SC-wLS~\cite{sc_wls_eccv2022} & 0.11, 0.7  & 0.42, 1.7 & 0.14, 0.6 & 0.39, 1.3 & 1.64, 0.9\\
     & NeuMap~\cite{neumap} & 0.06, \uline{0.25} & 0.19, 0.36 & 0.14, \uline{0.19} & 0.17, 0.53 & \textbf{0.06},  \uline{0.1}\\
     & \textbf{\ours}, K=20  & \uline{0.05}, 0.29 & \textbf{0.13}, \textbf{0.25} & 0.13, \textbf{0.18} & \textbf{0.06}, 0.22 & 0.12, \textbf{0.08}\\
     & \textbf{\ours-L}, K=20  & \uline{0.05}, 0.28 & \textbf{0.13}, \textbf{0.24} & \textbf{0.11}, \textbf{0.18} & \textbf{0.06}, \textbf{0.20} & 0.13, \textbf{0.08}\\
    \bottomrule 
    \end{tabular} 
    } \\[-0.25cm]
    \caption{\textbf{Comparison to the state of the art on Cambridge} with  the median translation (m) and angular ($^\circ$) errors.}
    \label{tab:cambridge}
    \vspace{-0.4cm} 
\end{table}

\subsection{Database compression}
\label{sub:compression}

\begin{figure}[h]
    \centering
    \includegraphics[width=\linewidth]{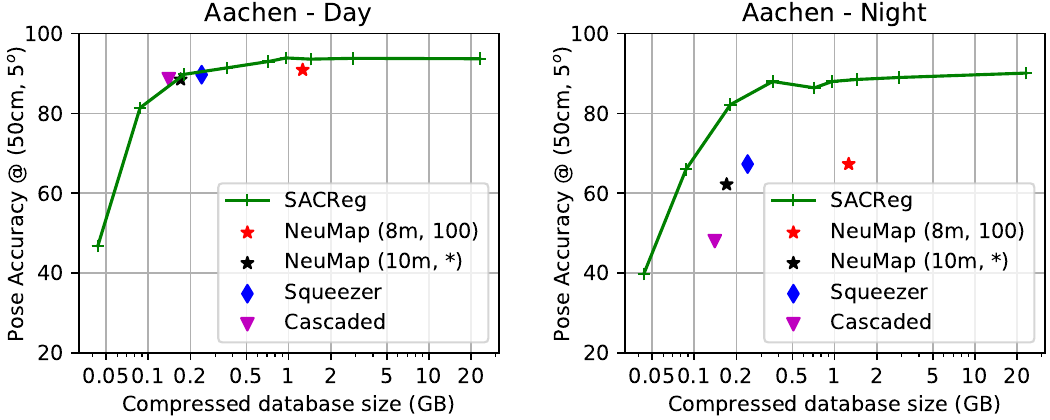}\\
    \vspace{-3mm}
    \caption{\textbf{Compression experiments.} We report the localization performance on Aachen as a function of the database storage size.
    \vspace{-0.4cm}
    }    
    \label{fig:xp_compr}
\end{figure}

One important limitation of the proposed method so far is the large volume of the pre-computed database image representations, if stored uncompressed.
Indeed, considering an input resolution $HW \triangleq 640 \times 480$, and a ViT-Base architecture with a patch size of 16px, an encoded image $\Raug_b$ requires $3.69\text{MB}$ of storage.
Product quantization (Section~\ref{sub:visloc}) 
allows a significant storage reduction
with negligible loss of performance.
We use a codebook of $256$ features per block, and vary the number of blocks for reaching different compression rates.

\PAR{Results.} 
During an offline phase, we compute, compress and store the representations of all database images.
At test time, we reconstruct the full token features from the stored codebook indices and the corresponding codebooks.
To alleviate the performance drop due to quantization, we slightly finetune the model for one additional epoch using compressed database features as inputs (considered as frozen), with a learning rate of $10^{-4}$ and a cosine-decay scheduler. This step is still scene-agnostic and is performed once for all.
In Figure~\ref{fig:xp_compr}, we report the performance on Aachen while varying the number of blocks of PQ quantization.
We observe that the performance remains similar with a compression factor up to 32, \ie, effectively reducing the database storage size from about 30GB (3.69MB/img) to 0.96GB (154kB/img).
Beyond this point, the performance gracefully degrades, such that for a compression factor of 128, our method is still able to obtain more than 80\% accuracy at 50cm\&5\textdegree{} on Aachen-Night.

\begin{table}
    \centering
    \resizebox{1\linewidth}{!}{
    \begin{tabular}{l|r|cc}
    \toprule
    Method & \multicolumn{1}{c|}{Size $\downarrow$} & Aachen-Day~$\uparrow$ & Aachen-Night~$\uparrow$ \\
    \midrule
    \textbf{\ours} (3.69MB/img) & 30.16  & \bf{85.3 / 93.7 / 99.6} & \bf{64.9 / 90.1 / 100.0} \\
    \midrule
    NeuMap (8m, 100)~\cite{neumap} & 1.26 & 80.8 / 90.9 / 95.6 & 48.0 / 67.3 / 87.8 \\
    \textbf{\ours}+PQ (154kB/img) & 0.96 & \bd{85.3} / \bd{93.9} / \bd{99.6} & \bd{62.8} / \bd{88.0} / \bd{100.0} \\
    \midrule
    \textbf{\ours}+PQ (58kB/img) & 0.36 & \bd{81.1} / \bd{91.4} / \bd{99.5} & \bd{59.7} / \bd{88.0} / \bd{100.0} \\
    Squeezer~\cite{yang2022scenesqueezer} & 0.24 & 75.5 / 89.7 / 96.2 & 50.0 / 67.3 / 78.6 \\
    \textbf{\ours}+PQ (29kB/img) & 0.18 & 76.6 / \bd{89.8} / \bd{98.9} & \bd{53.9} / \bd{82.2} / \bd{100.0} \\
    NeuMap (10m, *)~\cite{neumap} & 0.17 & 76.2 / 88.5 / 95.5 & 37.8 / 62.2 / 87.8 \\
    Cascaded~\cite{cheng2019cascaded} & 0.14 & \bd{76.7} / \bd{88.6} / 95.8  & 33.7 / 48.0 / 62.2 \\
    \textbf{\ours}+PQ (14kB/img) & 0.09 & \bd{61.8} / \bd{81.4} / \bd{98.2} & \bd{41.9} / \bd{66.0} / \bd{98.4} \\
     \bottomrule
    \end{tabular}
    }
    \vspace{-0.3cm}
    \caption{\textbf{Results with compression compared to the state of the art on Aachen Day-Night.} 
    The `Size' column represents the compressed dataset size in gigabytes.
    We highlight in \bd{bold} optimal values lying on an accuracy-versus-compression Pareto front.
    \vspace{-0.6cm}
    }
    \label{tab:quantizationsota}
\end{table}

\PAR{Comparison with the state of the art.}
In Figure~\ref{fig:xp_compr} and Table~\ref{tab:quantizationsota},
we compare our approach on the Aachen dataset with other scene-compression methods such as NeuMap~\cite{neumap}, which directly regress the 3D coordinates of a given set of 2D keypoints using learned neural codes, and other scene compression methods such as Cascaded~\cite{cheng2019cascaded} and Squeezer~\cite{yang2022scenesqueezer}, which are based on feature matching.  
Our approach achieves similar or better results compared to all other methods under similar compression ratios. 
Additionally, it is noteworthy to point out that, unlike NeuMap, we did not train our model on the Aachen dataset at all. 

\subsection{Scene coordinates regression}
\label{sub:coordinates_regression}

Lastly, to evaluate the regression performances of \ours{}, we apply our model on \emph{7-Scenes}, which provides dense ground-truth annotations. Using a shortlist size of $K=8$, we predict the 3D coordinates and corresponding confidence for each pixel of the test images.
We obtain a median and mean error of 4.2cm and 13.2cm respectively.
Results furthermore validate that confidence predictions are meaningful, as errors tends to get smaller when the confidence increases (Figure~\ref{fig:7-scenes}, top right). Confidence can thus be used as a proxy to filter out regions where errors are likely to be large (Figure~\ref{fig:7-scenes}, bottom right, and black regions in Figure~\ref{fig:regression_results}).

\section{Conclusion}

We introduce a novel paradigm for Scene Coordinates Regression with a model predicting pixelwise coordinates for a query image based on database images with sparse 2D-3D correspondences.
Our single model can be applied for visual localization in novel scenes of arbitrary scale without re-training, and outperforms other learning-based approaches that are trained for a single or a few small specific scenes.
Its database representations can be pre-computed offline for greater efficiency, and we furthermore show they can be highly compressed with negligible loss of visual localization performance.

{\small
\bibliographystyle{ieee_fullname}
\bibliography{biblio}
}

\clearpage

\appendix

\noindent{\huge\textbf{Appendix}\par}
\vspace{6mm}

\begin{figure*}
    \centering
    \resizebox{\linewidth}{!}{
    \begin{SCRfigtabular}
    \SCRfigline{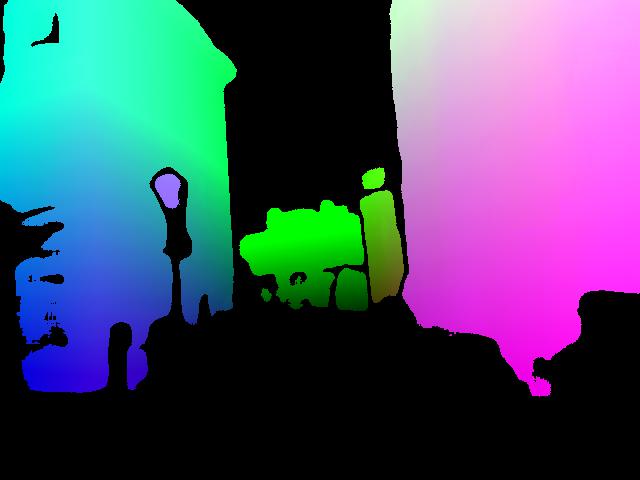} \\
    \SCRfigline{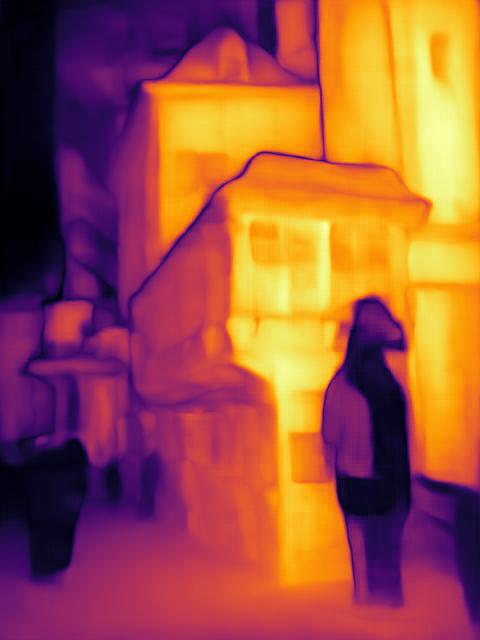} \\
    \SCRfigline{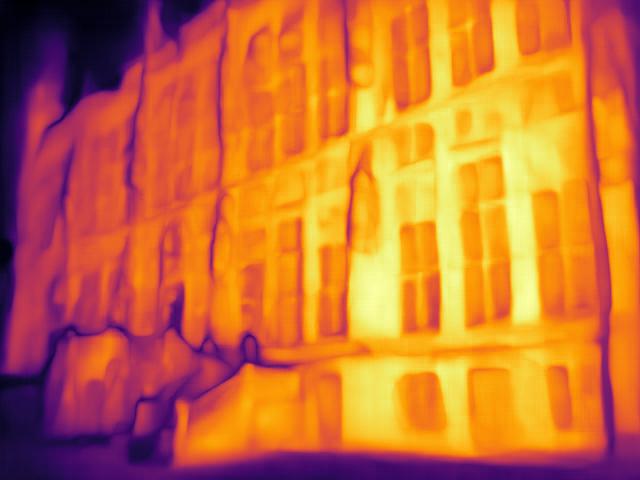} \\
    \SCRfigline{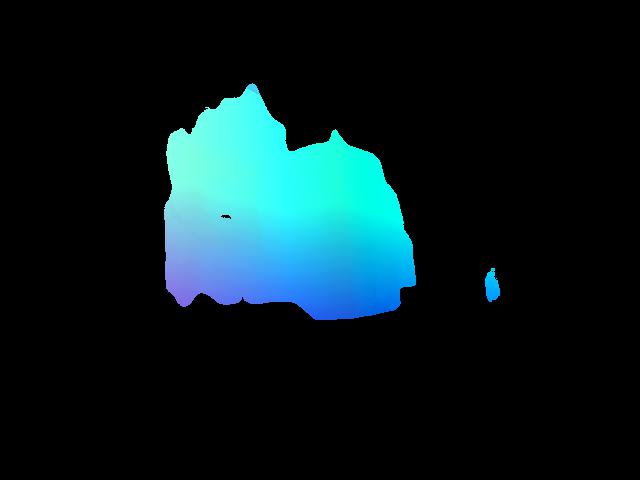} \\
    \SCRfigline{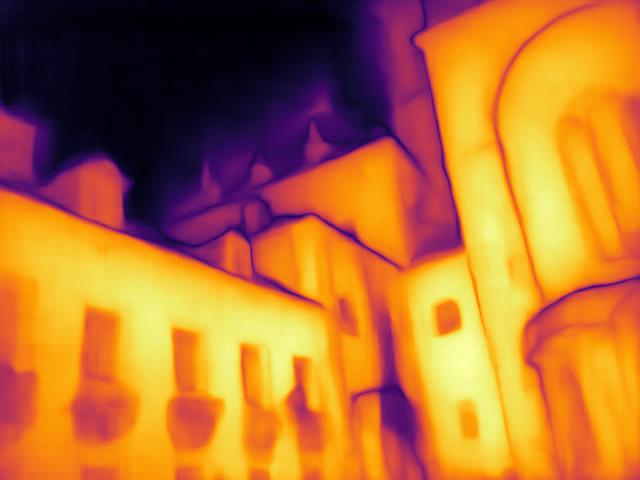} \\
    \SCRfigline{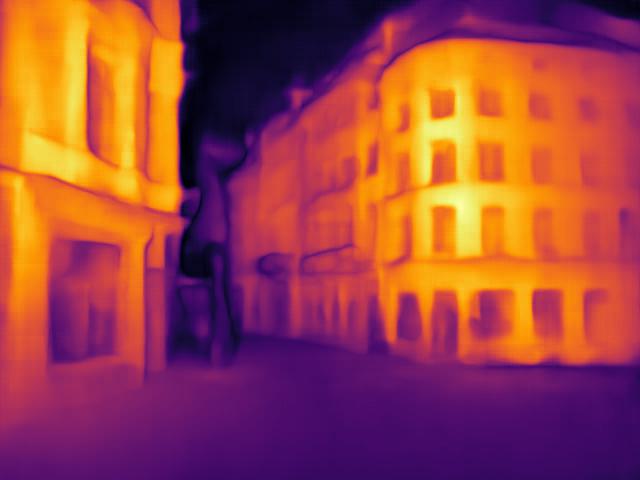} \\
    \end{SCRfigtabular}}
    \\[-0.2cm]
  \captionof{figure}{\textbf{Regression examples on Aachen-Day.}
    Our model predicts a dense 3D coordinates point map and a confidence map (\nth{5} and \nth{6} columns) for a given query image (\nth{4} column) using reference images retrieved from a \SfM{} database (\nth{1}, \nth{2}, \nth{3} columns). Only the first 3 reference images (out of $K=8$) are depicted.
    3D coordinates and confidence are colorized for visualization purposes, and areas with a confidence below $\tau=1$ are not displayed.
    Best viewed in color.
    }
    \label{fig:additional_regression_results}
\end{figure*}

\begin{figure*}[h]
\includegraphics[width=\linewidth]{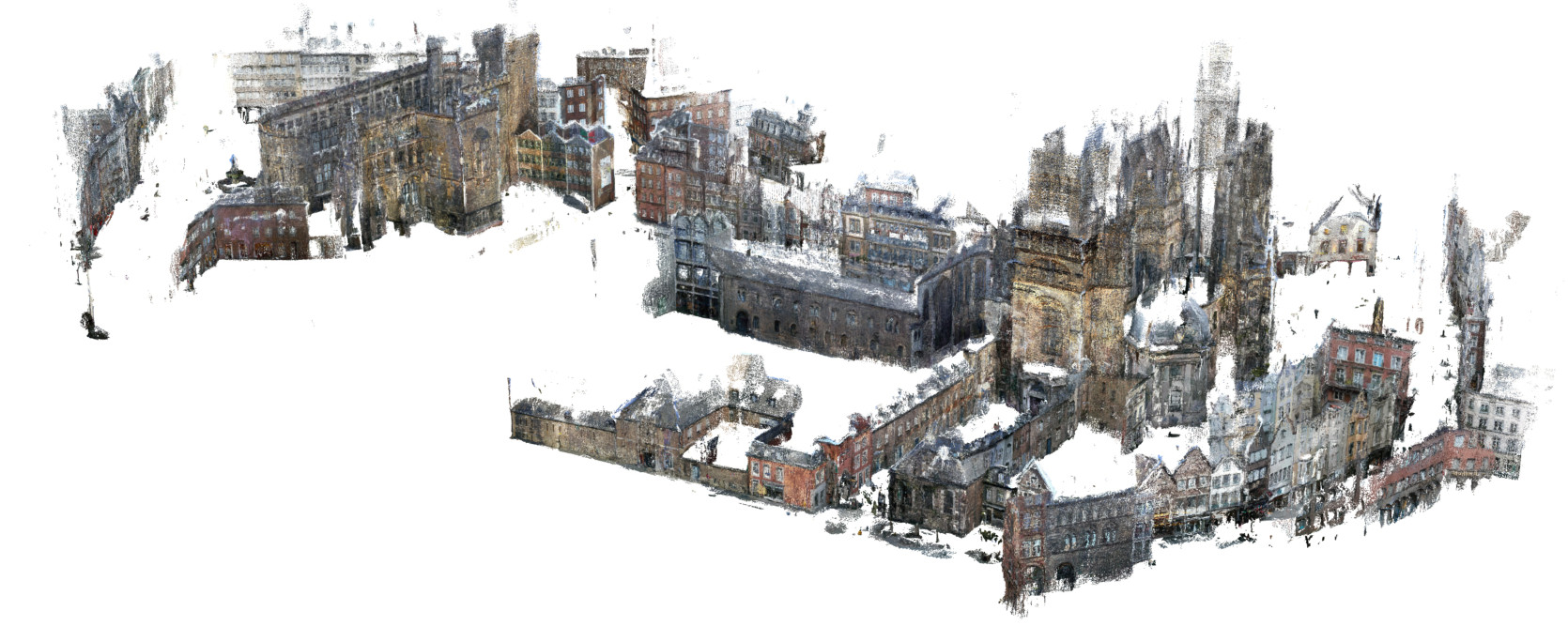} \\[-0.9cm]
\caption{\label{fig:large_scale_reconstruction} \textbf{Large-scale 3D reconstruction.} Dense point clouds regressed for different query images can be merged into a globally consistent 3D point cloud of a scene at scale. Example for query images from \emph{Aachen-Day}.}
\end{figure*}

In this appendix, we first present additional qualitative results in Section~\ref{sup:more}.
We then elaborate on the offline computation, compression and storage of database images %
 in Section~\ref{sup:compression}.
Next, we discuss architecture details of the 3D mixer (Section~\ref{sec:3dmixer}) and of the regression head (Section~\ref{sup:regression_head}).
We then provide detailed explanations and sketch proofs regarding the proposed point encoding and its inverse mapping in Section~\ref{sup:inverse}.
Finally, we %
give the hyper-parameters used for training the model in Section~\ref{sup:implem_details}.

\section{More qualitative results}
\label{sup:more}

Figure~\ref{fig:additional_regression_results} shows examples of regression outputs (dense 3D coordinates and confidence map) with their corresponding input images.
From these outputs, we can observe several key features of the model.
First, the model can successfully densify sparse inputs incoming from several images while respecting small structures like windows (\eg in rows \#3 and \#4).
Second, the model is robust to occlusions, as especially shown in rows \#2 and \#4 where occluded parts in the query \wrt database images are also rated as unreliable in the output confidence map.
Third, the model is typically not confident to extrapolate 3D predictions in query regions for which few or no 2D-3D annotations are available in the corresponding database images, such as the ground region, people or trees.

\PAR{3D reconstruction.}
The 3D coordinates predicted by \ours{} can be used to lift a query image into a 3D colored point cloud. 
Point clouds corresponding to different query images of the same scene can then be merged into a single large scale 3D reconstruction, as illustrated in Figure~\ref{fig:large_scale_reconstruction}.
In this case, we collected the output point clouds obtained for each query image from the Aachen-Day~\cite{aachen} dataset, removed low-confidence 3D points and simply concatenated all 3D points together before rendering the result using Blender~\cite{blender}.
Note that, because our approach directly predicts 3D coordinates in a metric world coordinate system, no further post-processing of the output is necessary to achieve these reconstructions.

\section{Database compression}
\label{sup:compression}

In this section, we give more details about the compression stage to store the pre-computed database representations, see last paragraph of Section 3.3 of the main paper.

\PAR{Raw storage.}
Considering an input resolution $HW \triangleq 640 \times 480$, and a ViT patch size of $p \triangleq 16$, an image representation typically consists of $H W/p^2=1200$ token features of $D \triangleq 768$ dimensions, each encoded using a 4-bytes floating point value. It thus leads to a memory use of $4 D H W/p^2 \approx 3.69\text{MB}$ per image.

\PAR{Product quantization (PQ)~\cite{jegou_pq_2011}}
is a lossy data compression technique that works as follows.
Given a vector $\x\in\mathbb{R}^D$ to compress, $\x$ is first split into $N$ sub-vectors denoted as \emph{blocks} $\x_j\in\mathbb{R}^{D/N}$, $j=1\dots N$, $D$ being a multiple of $N$, and the block size being $B=D/N$.
Each block $\x_j$ is then separately vector-quantized~\cite{vector_quantization} using a codebook $\C_j$.
Using codebooks with 256 entries, we can represent each block using a single byte index, effectively decreasing the memory requirement by a factor of $4B$ (assuming 4 bytes requirements per floating point value).
In our specific case, given a 3D-augmented database image representation $\Raug=\{\x_i\}$ composed of token features $\x_i\in\mathbb{R}^D$, we split each token features into $N$ blocks $\x_{i,j}\in\mathbb{R}^B$ and store for each one the index of the closest entry from the codebook $\C_j$. %
The original feature can be approximately reconstructed by the codebook entry corresponding of the stored index for each block.

\PAR{Codebook learning.}
To learn the set of codebooks $\{\C_j\}$, we first select a set of 30,000 database images $\{\I_k\}$ from our %
training set, equally sampled from the \emph{Habitat}, \emph{ARKitScenes} and \emph{MegaDepth} datasets.
We encode them,  along with their associated 2D-3D annotations, and get their 3D-augmented representation $\{\Raug_k\}$, each of which being a set of 1200 token features for a resolution of $640\times 480$. 
Gathering tokens from all images results in a set of $M$ features vectors $\x_i \in \mathbb{R}^{D}, i=1,\dots,M$.
We split each vectors into $N$ blocks $(\x_{i,1}, \dots, \x_{i,N})$ of dimensions $B$, and we cluster features $(\x_{i,j})_{i=1,\dots,M}$ for each block $j\in\{1,\dots,N\}$ into $k=256$ centroids using $k$-means~\cite{kmeans} to form the codebook $\C_j$.

\PAR{Results.} 
Localization results after quantization and without finetuning are reported in Table~\ref{tab:quantization}.
For 7-Scenes and Cambridge-Landmarks, we report the average median translation error over \{Chess, Fire, Heads, Office	Pumpkin, Kitchen, Stairs\} for 7-Scenes and \{ShopFacade, OldHospital, College, Church, Court\} for Cambridge-Landmarks.

We first observe that a quantization with a block size $B=2$ or $B=4$ has a limited impact on localization accuracy, but that performance progressively degrades when using more aggressive compression schemes.

We also try finetuning the model for one additional epoch on the training set using compressed database features as inputs (without backpropagating to them, \ie, they are considered as fixed), with a learning rate of $10^{-4}$ and a cosine-decay scheduler. 
We observe large performance improvements thanks to this fine-tuning: the models compressed with $B=6$ or $B=8$ achieve performances after fine-tuning similar to the original one, while requiring only 115kB of storage per image.
These data points are the ones plotted in Figure 6 of the main paper and reported in Table 7 of the main paper.

\begin{table*}
    \centering
    \resizebox{0.7\linewidth}{!}{
    \begin{tabular}{l|c|c|cccc}
    \toprule
    Quantization & FT & kB/img $\downarrow$ &  Camb.~$\downarrow$ & 7scenes~$\downarrow$ & Aachen-Day~$\uparrow$ & Aachen-Night~$\uparrow$ \\
    \midrule
    Raw & & 3686 & 0.0976 & 0.036428571 & 85.3 / 93.7 / 99.6 & 64.9 / 90.1 / 100.0 \\
    \midrule
    PQ (B=2) & & 461 & 0.0998 & 0.036857143 & 86.2 / 94.1 / 99.6 & 63.9 / 89.5 / 100.0 \\
    PQ (B=4) & & 230 & 0.1014 & 0.037857143 & 84.7 / 94.1 / 99.6 & 65.4 / 89.0 / 100.0   \\
    PQ (B=6) & & 154 & 0.1104 & 0.039857143 & 83.0 / 93.1 / 99.6 & 60.2 / 83.2 / 100.0 \\
    PQ (B=8) & & 115 & 0.125 & 0.042571429 & 80.2 / 91.4 / 99.5 & 56.5 / 80.6 / 100.0 \\
    PQ (B=16) & & 58 & 0.2526 & 0.075857143 & 38.7 / 62.6 / 86.5 & 7.9 / 17.8 / 45.0 \\
    PQ (B=32) &  & 29 & 26.3834 & 1.674 & 0.0 / 0.0 / 0.7 & 0.0 / 0.0 / 0.0 \\
    PQ (B=64) & & 14 & 33.3698 & 1.702 & 0.0 / 0.0 / 0.0 & 0.0 / 0.0 / 0.0 \\
    PQ (B=128) & & 7 & 50.2998 & 1.836142857 & 0.0 / 0.0 / 0.0 & 0.0 / 0.0 / 0.0 \\
    \midrule
    PQ (B=2) & \checkmark & 461 & 0.0988 & 0.036142857 & 85.0 / 93.8 / 99.6 &  66.0 / 89.0 / 100.0 \\
    PQ (B=4) & \checkmark & 230 & 0.0992 & 0.037 & 85.7 / 93.6 / 99.5 &  62.8 / 88.5 / 100.0 \\
    PQ (B=6) & \checkmark & 154 & 0.1004 & 0.037142857 & 85.3 / 93.9 / 99.6 & 62.8 / 88.0 / 100.0 \\
    PQ (B=8) & \checkmark & 115 & 0.1072 & 0.037428571 & 85.1 / 93.0 / 99.5 & 63.9 / 86.4 / 100.0 \\
    PQ (B=16) & \checkmark & 58 & 0.117 & 0.042285714 & 81.1 / 91.4 / 99.5 & 59.7 / 88.0 / 100.0  \\
    PQ (B=32) & \checkmark & 29 & 0.1456 & 0.047 & 76.6 / 89.8 / 98.9 & 53.9 / 82.2 / 100.0 \\
    PQ (B=64) & \checkmark & 14 & 0.2302 & 0.059142857 & 61.8 / 81.4 / 98.2 & 41.9 / 66.0 / 98.4 \\
    PQ (B=128) & \checkmark & 7 & 1.4034 & 0.089714286 & 21.7 / 46.8 / 92.2 & 13.6 / 39.8 / 89.0 \\
     \bottomrule
    \end{tabular}
    }
    \vspace{-0.3cm}
    \caption{\textbf{Product quantization (PQ)~\cite{jegou_pq_2011}} enables to compress database tokens with minimal performance losses.
    Results obtained for $640 \times 480$ resolution images 
     using K=20 retrieved database images, for different sizes of quantization blocks $B$. 
    The compression factor with respect to storing raw representations is $4B$. 
    kB/img represents the number of kilobytes required to store 
    the compressed  %
    representations of a single database image.
    FT denotes where we perform one epoch of fine-tuning using the fixed compressed database representations. 
    }
    \label{tab:quantization}
\end{table*}

\section{Architecture of the 3D mixer}
\label{sec:3dmixer}

In this section, we give more details about the architecture of the 3D mixer, which relies on a transformer decoder to enrich the image tokens from the database images with the information from the sparse 2D-3D annotations.

\begin{figure}
    \centering
    \includegraphics[trim=0 350 390 0,clip,width=\linewidth]{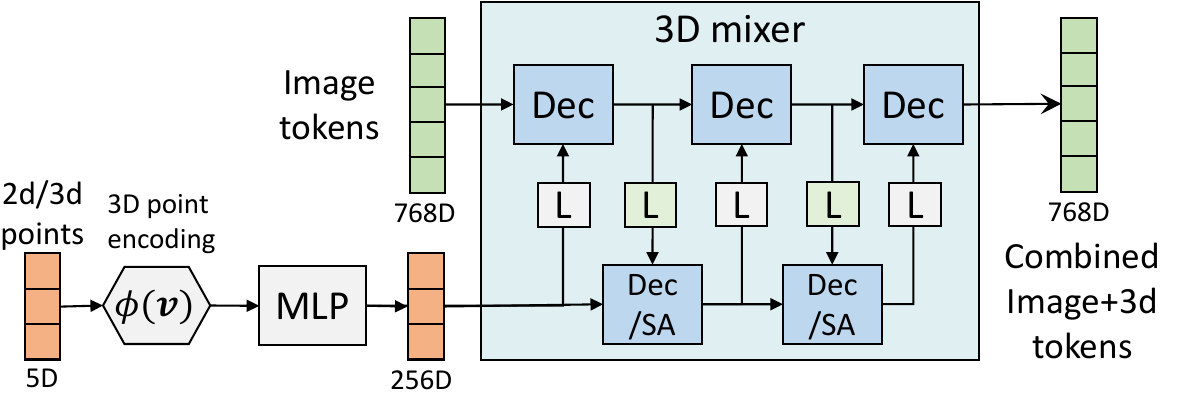} \\[-0.25cm]
    \caption{\textbf{Overview of the 3D-mixer} which combines together the image with its associated 2D-3D sprase correspondences in a single representation. ``Dec'' stands for a classical transformer decoder block with self-attention (SA), cross-attention (CA) and MLP. ``Dec /SA'' is a decoder without self-attention to scale to large number of 3D points. The linear projection (``L'') enables to map between the dimensions of image and 3D point tokens (768 \vs 256).}
    \label{fig:3dmixer}
\end{figure}

Figure~\ref{fig:3dmixer} gives on overview of the architecture of the 3D mixer. 
We first embed each sparse 2D-3D correspondence in a 256-dimensional feature space, denoting the result as \emph{point tokens}.
This is performed by first encoding the 3D coordinates using $\phi$ as described in Section 3.2 of the main paper and by feeding these to an MLP.
A series of transformer decoder blocks, where each block consists of a multi-head self-attention~(SA) between the $768$-dimensional image tokens, 
a multi-head cross-attention~(CA) that shares information from all point tokens with these image tokens, and an MLP then allow to enrich the image tokens with 2D-3D points information.

For robustness against potential outliers in the input sparse 3D points (due to \eg triangulation noise), we interleave the image-level decoder blocks with a new set of point-level decoder blocks, improving capability of the model to focus attention on specific point tokens.
We remove self-attention in these point-level decoder blocks to reduce their complexity. 
To account for the different features sizes between point tokens ($256$) and image tokens ($768$), we use linear projections that are shared across decoder blocks, see Figure~\ref{fig:3dmixer}.
In practice, we use $4$ image-level decoder blocks 
interleaved with 3 point-level decoder blocks.
We now evaluate the design choices of the 3D mixer architecture.

\begin{table}[]
    \centering
    \resizebox{\linewidth}{!}{
    \begin{tabular}{l|ccc}
    \toprule
    3D mixer architecture & Camb.~$\downarrow$ & 7scenes~$\downarrow$ & Aachen-Night~$\uparrow$\\
    \midrule 
    Simple, depth=1 & 0.66 & 0.11 & 16.8 / 40.8 / 87.4 \\
    Simple, depth=2 & 0.61 & \textbf{0.08} & 20.4 / 42.4 / 88.0 \\
    Simple, depth=4 & 0.55 & 0.10 & \textbf{22.0} / 39.3 / 89.5 \\
    \midrule 
    Alternating, depth=1 & 0.56 & 0.11 &	18.8 / 41.4 / 88.0 \\
    Alternating, depth=2 & 0.46 & 0.11 &	\textbf{22.0} / 46.1 / 88.5 \\
    \hl Alternating, depth=4 & \hl \textbf{0.43} & \hl 0.11 & \hl \textbf{22.0} / \textbf{47.1} / \textbf{90.6} \\
    \bottomrule
    \end{tabular}
    }\\[-0.25cm]
    \caption{\textbf{Ablation on the 3D mixer architecture. `Simple' denote image-level decoder blocks while `Alternating' refer to interleaved image-level and point-level decoder blocks.}
        }
    \label{tab:ab_mixer3d}
\end{table}

\PAR{Ablations on the 3D mixer architecture.}
We report results for a few variants of the 3D mixer architecture in Table~\ref{tab:ab_mixer3d}.
We compare the interleaved decoder architecture from Figure~\ref{fig:3dmixer} with a simpler architecture composed of a sequence of several image-level decoder blocks (`Simple').
In all cases, increasing the number of decoder blocks leads to improved performance.
However, we observe a substantial gain with the alternating decoder architecture, thanks to the enhanced denoising ability of this architecture.

\section{Regression head architecture}
\label{sup:regression_head}

\begin{figure*}
    \centering
    \includegraphics[trim=0 300 300 0,clip,width=0.7\linewidth]{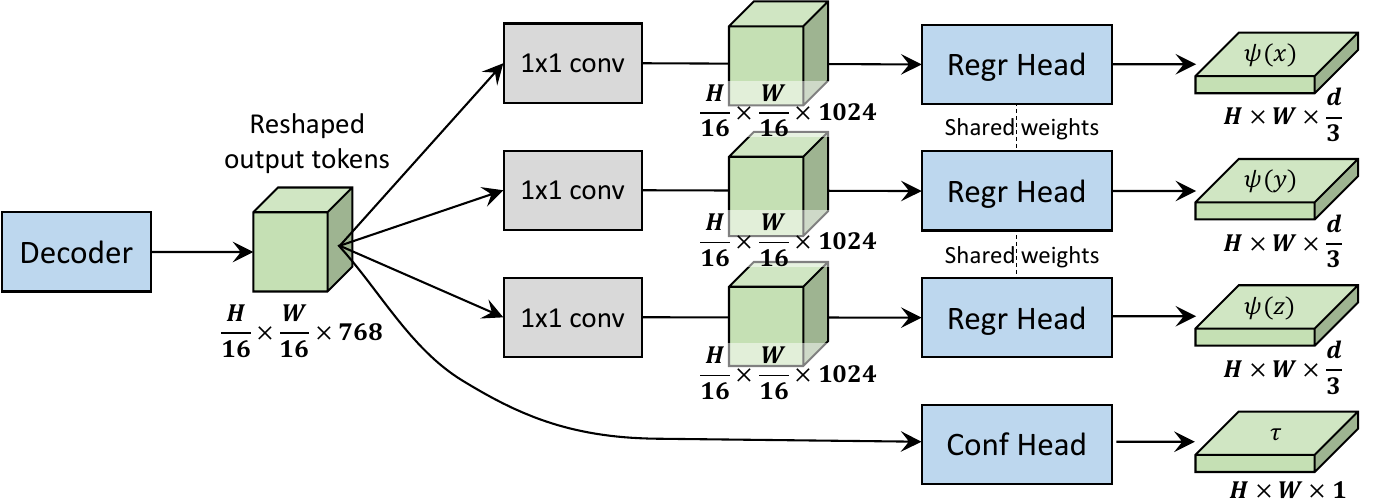} \\[-0.25cm]
    \caption{\bd{Detailed pipeline of the decoupled regression head.}
            After reshaping the output tokens from the decoder into a dense feature map, 3 distinct $1 \times 1$ convolutions extract channel-wise features from the same feature map. Then, a regression head with shared weight is applied to regress 3D point encodings for each channel.
            In parallel, another head takes care of predicting the confidence from the initial feature map.
            }
    \label{fig:head1}
\end{figure*}

We now give more details about the regression head.
In Section~4.2 of the main paper, we experiment with different architectures and finally retain the one that regresses each of the 4 $x,y,z,\tau$ channels separately.
In this case, we share the prediction head between all spatial x, y and z components as shown in Figure~\ref{fig:head1}, and another head serves to predict the pixel-wise confidence.

\begin{figure*}
    \centering
    \includegraphics[trim=0 400 60 0,clip,width=0.8\linewidth]{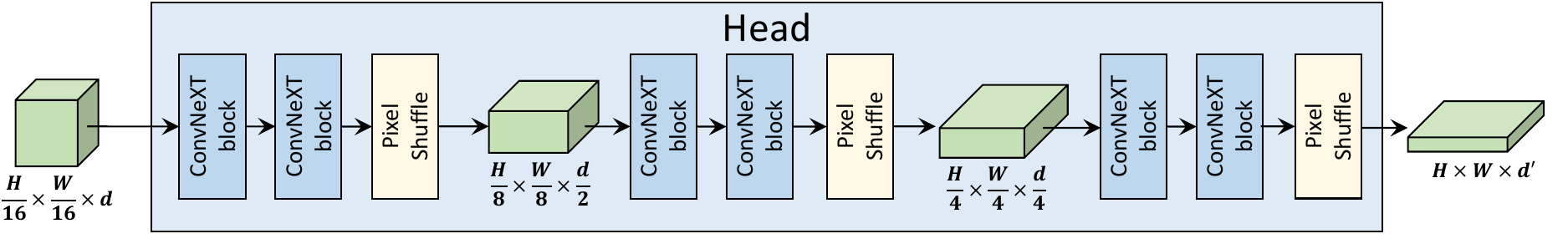} \\[-0.25cm]
    \caption{\bd{Internal architecture of the regression head}.
            It consists of a sequence of 6 ConvNeXT~\cite{liu2022convnet} blocks interleaved with 3 PixelShuffle~\cite{pixelshuffle_superresol} operations to increase the resolution while simultaneously halving the channel dimension.
            In this figure, we denote the input feature dimension as $d$ and the output dimension as $d'$.
            }
    \label{fig:head2}
\end{figure*}

All heads (\ie, 3D regression heads and confidence head) share the same architecture, as specified in the main paper and illustrated in Figure~\ref{fig:head2}.
Specifically, the head consists of a sequence of 6 ConvNeXT~\cite{liu2022convnet} blocks interleaved with 3 PixelShuffle~\cite{pixelshuffle_superresol} operations to increase the resolution while simultaneously halving the channel dimension.

\section{Point encoding}
\label{sup:inverse}
We provide in this section further information regarding the point encoding $\phi$ presented in the paper.
First recall that we have defined $\phi(\boldsymbol{v})=\left[\psi(x),\psi(y),\psi(z)\right]$
for $\bm{v}=(x,y,z) \in \mathbb{R}^3$, \ie, simply concatenating the encodings of the $x$,
$y$ and $z$ channels, by a mapping $\psi$.
For the sake of brevity, most of the discussion will thus focus on this mapping:
\begin{equation}
\begin{aligned}
\psi: t \in \mathbb{R} \rightarrow &\left[ \cos(f_{1}x),\sin(f_{1}x), \dots,\cos(f_{F}x),\sin(f_{F}x) \right] \\
&\in (\mathbb{S}^1)^F \subset [-1,1]^{2F},
\end{aligned}
\end{equation}
defined for some positive frequencies $\left\{ f_{i}\right\}_{i=1\ldots F}$, with
$F=d/6\in\mathbb{N}^*$, where $d$ is the dimension of the output point encoding $\phi(\bd{v})$.
We furthermore typically consider harmonically distributed frequencies $f_{i}=f_{1}\gamma^{i-1}$ for $i=1\ldots F$, where $\gamma > 1$.

\subsection{Injectivity / periodicity}

\PAR{Lemma 1.}
$\gamma \in \mathbb{R}\setminus\mathbb{Q} \Rightarrow \psi(x)$ is injective, or equivalently, if $\gamma$ is irrational then there is no $x$ and $y$ with $x\neq y$ such that $\psi(x)=\psi(y)$.

\PAR{Proof by contraposition.}
We will prove that $\psi(x)=\psi(y)\text{ with }x\neq y \Rightarrow \gamma \in \mathbb{Q}$.
For simplicity and without loss of generality, assume $F=2$ frequencies.
Let $x,y \in \mathbb{R}$ two distinct numbers ($x \neq y$).
The proposition $\psi(x)=\psi(y)$ is equivalent to:
\begin{equation}
\begin{cases}
\begin{aligned}
\cos(f_{1}x) & = \cos(f_{1}y) \\
\sin(f_{1}x) & = \sin(f_{1}y) \\
\cos(f_{2}x) & = \cos(f_{2}y) \\
\sin(f_{2}x) & = \sin(f_{2}y),
\end{aligned}
\end{cases}
\end{equation}

which can be rewritten as: 
\begin{equation}
\begin{cases}
\begin{aligned}
e^{if_{1}x} & = e^{if_{1}y}\\
e^{if_{2}x} & = e^{if_{2}y},
\end{aligned}
\end{cases}
\end{equation}

\ie, that there exist $k_{1},k_{2}\in\mathbb{Z}^{*}$ such that:
\begin{equation}
\begin{cases}
\begin{aligned}
f_{1}x & = f_{1}y+2\pi k_{1} \\
f_{2}x & = f_{2}y+2\pi k_{2}. 
\end{aligned}
\end{cases}
\end{equation}
Rearranging the equations, we get:
\begin{numcases}{}
\begin{aligned}
\frac{f_{1}}{2\pi}(x-y) & = k_{1} \label{eq:f1}\\
\frac{f_{2}}{2\pi}(x-y) & = k_{2}. \label{eq:f2}
\end{aligned}
\end{numcases}
Dividing the second term by the first one, which is possible as 
$k_{1},k_{2}\neq0$, yields:
\begin{equation}
\frac{f_{2}}{f_{1}}=\gamma=\frac{k_{2}}{k_{1}}\in\mathbb{Q} \quad\blacksquare
\end{equation}

While any irrational $\gamma$ fulfils the injectivity lemma, we
note that algebraic numbers should ideally be avoided. Indeed they
introduce a risk that $\psi$ would be periodic over a subset of dimensions
$\xi \subseteq \{1,\ldots,F\}$ , with $\left|\xi\right|>1$. 
Take, for instance, an algebraic $\gamma=\sqrt{2}$ and assume $F=3$ frequencies. 
We then have $f_{2}=\sqrt{2}f_{1}$ and $f_{3}=2f_{1}$.
In this case, one can trivially show that $\psi_\xi = [\psi_i]_{i\in\xi}$ is periodic for $\xi=\{1,3\}$.
In the presence of noise, this can severely impair the robustness of the inverse mapping. 
Hence, in general, we recommend choosing transcendental values for $\gamma$, which is not really a problem as \emph{almost all real }numbers are transcendental, \ie, $p(X\text{ is transcendental})=1$ for a uniformly random real $X$ in any given range $[a,b]$ with $a<b$.

\paragraph{Numerical considerations.}
Modern representations of real numbers on computers are limited by hardware constraints. 
In practice, a real number $x\in\mathbb{R}$ is stored as a fixed-length bit vector (64 bits in our case), 
which roughly amounts to storing $x$ as a rational $\frac{a}{b},$ with $a,b\in\mathbb{Z}^*$.
Let then denote all frequencies as rational numbers $f_i=\frac{a_i}{b_i}$, with coprime $a_i,b_i\in\mathbb{N}^*$.
In this case, $\psi$ is periodic, %
with a period $P \neq 0$ multiple of all individual periods $2 \pi / f_i$, for $i=1, \ldots, F$. 
Equivalently, there exist some non-zero integers $k_i \in \mathbb{N}^*$ such that:
\begin{equation}
P = k_{i} \cfrac{2 \pi}{f_i}, \quad \forall i=1 \ldots F
\end{equation}
\ie:
\begin{equation}
P = 2 \pi k_{i} \cfrac{b_i}{a_i}, \quad \forall i=1 \ldots F.
\end{equation}
The smallest period $P$ satisfying all these $F$ equations is thus $P=2 \pi ~ \texttt{lcm}\{b_i\} / \texttt{gcd}\{a_i\}$, defined by the least common multiple of all denominators ($\texttt{lcm}$) divided by the greatest common divisor ($\texttt{gcd}$) of all numerators. Such ratio is in general hard to predict for arbitrary frequencies $\{f_i\}$, %
but it can be extremely large when considering irreducible fractions of large terms $a_i,b_i \simeq 10^{15}$.
For our use-case, we want the period to be a large as possible, and in practice, using double floating-point representations, we can find values for $\gamma$ yielding periods in the order of billions of kilometers, which is more than sufficient.

\begin{table*}
    \centering
    \begin{tabular}{cc|cc|cc}
    \toprule 
    $F$ & $d{=}6F$ & $f_{1}$ & $\gamma$ & $P_{1}$ & $P_F$\\
    \midrule
    4 & 24 & 0.020772487794205544 & 5.7561020938998690 & 302m & 1.59m \\
    6 & 36 & 0.017903170262351338 & 3.7079736887249526 & 351m & 50cm \\
    8 & 48 & 0.031278470093268460 & 2.5735254599557535 & 201m & 27cm \\
    \bottomrule 
    \end{tabular}
    \vspace{-0.3cm}
    \caption{\textbf{%
    Frequency parameters} $f_1$ and $\gamma$ used in our experiments.
            $P_1$ and $P_F$ are respectively the periods of the lowest and highest frequencies.}
    \label{tab:freqs}
\end{table*}

\paragraph{Practical choice.}

We thus adopt an empirical stance and evaluate several random values
for the frequency-generator parameters $f_{1}$ and $\gamma$. 
In particular, we impose that they satisfy the metric scale conditions, \ie, the smallest and largest periods must lie in the following metric ranges: 
\begin{eqnarray}
    P_1 = \frac{2\pi}{f_{1}}\in [200\text{m},500\text{m}] \\
    P_F = \frac{2\pi}{f_F}  \in [20\text{cm},2\text{m}]. \nonumber 
\end{eqnarray}
Note that the 200-500 meters range specified for $P_{1}$ does not relate to the scale of the full dataset, but to the scale of the point cloud that can be observed through a single query view. 
In the unlikely case that this observable point cloud is larger than 200 meters, our theoretical results above still guarantees that any encoding admits a unique pre-image in the absence of noise. 
In the presence of noise, however, it could become more likely that large errors would happen during the inverse mapping.

To select the best frequency parameters, we train the network on $224{\times}224$ images on a subset of 20K training pairs for 20 epochs and choose the parameters that yield the maximum accuracy over a small validation set of the same size. We give in Table~\ref{tab:freqs} the resulting values that we have used in our experiments for different number of frequencies $F$.

\subsection{Inverse projection}
We formulate the inverse mapping problem as a non-linear least-square projection:
\begin{equation}
\label{eq:inv_def}
\psi^{-1}(\boldsymbol{y})=\text{\ensuremath{\arg\min_{t\in B}}}\left\Vert \psi(t)-\boldsymbol{y}\right\Vert ^{2},
\end{equation}
where $B$ is a search domain constructed as the union of point clouds
ranges for each input database image $I_{k}$. For instance, and without
loss of generality, when considering the $x$ channel of some 3D coordinates, $B$ is formally
defined 
\begin{equation}
B=\bigcup_{k=1\ldots K}\left[\min\left\{ x_{j}^{k}\right\} ,\max\left\{ x_{j}^{k}\right\} \right],
\end{equation}
where the set $\{x_{j}^{k}\}$ belongs to the sparse 2D-3D annotation
set $\mathcal{V}_{k}=\left\{ (p_{j}^{k},v_{j}^{k})\right\} $ associated
with each database image $\mathcal{I}_{k}$.
While $B$ should ideally be the infinite real space in theory, we found important in practice to narrow down the search range for efficiency reasons as well as noise robustness.

The minimization objective of Equation~\ref{eq:inv_def} can be expanded:
{
\small
\begin{equation}
\begin{aligned}
S(t) & =\left\Vert \psi(t)-\boldsymbol{y}\right\Vert ^{2}\\
 & =\sum_{i=1\ldots F} \big[ \left(\cos(f_{i}t)-\boldsymbol{y}_{2i-1}\right)^{2} +\left(\sin(f_{i}t)-\boldsymbol{y}_{2i}\right)^{2} \big]\\
 & =\sum_{i=1\ldots F} \big[ \cos^{2}(f_{i}t)-2\boldsymbol{y}_{2i-1}\cos(f_{i}t)+\boldsymbol{y}_{2i-1}^{2} \\
 & \quad\quad\quad\quad\quad\quad\quad+\sin^{2}(f_{i}t)-2\boldsymbol{y}_{2i}\sin(f_{i}t)+\boldsymbol{y}_{2i}^{2} \big]\\
 & =F+\left\Vert \boldsymbol{y}\right\Vert ^{2}-2\sum_{i=1\ldots F} \big[ \boldsymbol{y}_{2i-1}\cos(f_{i}t)+\boldsymbol{y}_{2i}\sin(f_{i}t) \big].
 \label{eq:sum1}
 \end{aligned}
 \end{equation}
 }

Using the trigonometric identity:
\begin{equation}
A\cos(x)+B\sin(x)=\sqrt{A^{2}+B^{2}}\cos\left(x-\arctan\frac{B}{A}\right)
\end{equation}
with the convention $\arctan( \pm \infty) = \pm \pi/2$,
we can simplify Equation~\ref{eq:sum1} into:
\begin{align}
S(t) & = F + \|\bm{y} \|^2 -2\sum_{i=1\ldots F} \sqrt{\boldsymbol{y}_{2i-1}^2 + \boldsymbol{y}_{2i}^2}\nonumber\\
    &  \quad\quad\quad\quad\quad\quad\quad\quad \times\cos\left(f_{i}t-\arctan\frac{\boldsymbol{y}_{2i}}{\boldsymbol{y}_{2i-1}}\right).\nonumber\\
\label{eq:sumcos}
\end{align}

In other words, the inverse projection consists in finding the minimum of a weighted sum of cosines with different frequencies.
This problem has no analytic solution in general for $F>2$. It can however be approximately solved when considering inputs normalized by pairs (\ie $\bm{y} \in (\mathbb{S}^1)^F$, or $\boldsymbol{y}_{2i-1}^2 + \boldsymbol{y}_{2i}^2$=1).

Inspired by~\cite{cosine}, we efficiently probe the search space at regular intervals $t_{n}=\frac{n\pi}{2f_{F}}\in B$, with $n\in\mathbb{Z}$ a signed integer and where $f_{f}$ is the highest frequency.
A reasoning similar to the one developed in~\cite{cosine} suggests that at most one minima can lie in $]t_{n},t_{n+1}[$, in which case we have $S'(t_{n})<0$ and $S'(t_{n+1})>0$, denoting $S'$ the derivative of $S$.
The location of the corresponding minima can be approximated as:
\begin{equation}
t=t_{n}-\left(t_{n+1}-t_{n}\right)\frac{S'(t_{n},\boldsymbol{y})}{S'(t_{n+1},\boldsymbol{y})-S'(t_{n},\boldsymbol{y})}.
\end{equation}

Finally, we return the local minimum approximation $t$ for which $S(t)$ is smallest over all $t_{n}$, which in practice is a good estimate of the global minimum.
Typically, for $f_{F}\simeq10 m^{-1}$, this amounts to probe every $\simeq10$ centimeters, which translates into 200 evaluation of $S$ per channel and per pixel for a typical 20-meter-wide outdoor scene.

\section{Training hyperparameters}
\label{sup:implem_details}

We report the detailed hyperparameter settings we use for training SACReg in Table~\ref{tab:training_step234}.

\begin{table*}[]
    \centering
    \begin{tabular}{l@{\hskip 1.0cm}l@{\hskip 0.8cm}l}%
    \toprule
    Hyperparameters & low-resolution training & high-resolution training \\
    \midrule
    Optimizer & AdamW~\cite{adamw} & AdamW~\cite{adamw} \\
    Base learning rate & 1e-4 & 1e-4 \\
    Weight decay & 0.05 & 0.05 \\
    Adam $\beta$ & (0.9, 0.95) & (0.9, 0.95) \\
    Batch size & 128 & 128 \\
    \multirow{2}{*}{Epochs / Learning rate scheduler} & \multirow{2}{*}{100 epochs with a fixed lr} & 20 epochs with fixed lr \\ & &~+20 epochs with cosine decay \\
    Warmup epochs & 10 & 2 \\
    \midrule
    Frozen modules & Encoder & - \\
    \midrule
    Number of database views & 1 & 1 \\
    Input resolution & $224{\times}224$ & $512{\times}384$ 3 \\
    Image Augmentations & Random crop, color jitter & Random crop, color jitter  \\
    Sampled 3D points & 1024 points per DB image & 1024 points per DB image \\
    \cmidrule{2-3}%
    \multirow{4}{*}{3D Augmentations} & 5\% triangulation noise & 5\% triangulation noise \\
                                      & rotations, translations in & rotations, translations in \\
                                      & $[-1000,1000]^3$, rescale & $[-1000,1000]^3$, rescale  \\
                                      & in $[0.5,2]$ & in $[0.5,2]$ \\
    \cmidrule{2-3}%
    \multirow{2}{*}{Retrieval Augmentations} & Gaussian noise (std=0.05) & Gaussian noise (std=0.05) \\
                                     & on FiRE retrieval scores & on FiRE retrieval scores \\
    \bottomrule
    \end{tabular}
    \vspace{-0.3cm}
    \caption{\textbf{Detailed hyper-parameters} for the training, with first a low-resolution training with frozen encoder before the higher-resolution training, in order to save training time} %
    \label{tab:training_step234}
    \vspace{1.7cm}
\end{table*}

\end{document}